\documentclass[conference]{IEEEtran}
\IEEEoverridecommandlockouts

\usepackage{wrapfig}

\usepackage{algorithmic}
\usepackage{graphicx}
\usepackage{textcomp}
\usepackage{xcolor}
\usepackage{color}
\usepackage{algorithm}
\usepackage{booktabs}
\usepackage{comment}
\usepackage{pifont}
\usepackage{multirow}
\usepackage{subfigure}
\usepackage{gensymb}

\newcommand{\eat}[1]{}

\usepackage{cite}
\usepackage{amsmath,amssymb,amsfonts}
\usepackage{algorithmic}
\usepackage{hyperref}  
\usepackage{graphicx}
\usepackage{textcomp}
\usepackage{xcolor}
\def\BibTeX{{\rm B\kern-.05em{\sc i\kern-.025em b}\kern-.08em
    T\kern-.1667em\lower.7ex\hbox{E}\kern-.125emX}}
\begin{document}


\title{Accelerate Coastal Ocean Circulation Model with AI Surrogate}

\author{
    \IEEEauthorblockN{
        Zelin Xu\IEEEauthorrefmark{1}, 
        Jie Ren\IEEEauthorrefmark{2}, 
        Yupu Zhang\IEEEauthorrefmark{1}, 
        Jose Maria Gonzalez Ondina\IEEEauthorrefmark{3}, 
        Maitane Olabarrieta\IEEEauthorrefmark{3}, \\
        Tingsong Xiao\IEEEauthorrefmark{1}, 
        Wenchong He\IEEEauthorrefmark{1}, 
        Zibo Liu\IEEEauthorrefmark{1}, 
        Shigang Chen\IEEEauthorrefmark{1}, 
        Kaleb Smith\IEEEauthorrefmark{4}, 
        Zhe Jiang\IEEEauthorrefmark{1}\IEEEauthorrefmark{5}\thanks{\IEEEauthorrefmark{5} Corresponding author.}
    }
    
    \IEEEauthorblockA{\IEEEauthorrefmark{1}Department of Computer \& Information Science \& Engineering, University of Florida, Gainesville, FL, USA}
    
    \IEEEauthorblockA{\IEEEauthorrefmark{2}Department of Computer Science, College of William \& Mary, Williamsburg, VA, USA}
    
    \IEEEauthorblockA{\IEEEauthorrefmark{3}Department of Civil \& Coastal Engineering, University of Florida, Gainesville, FL, USA}
    
    \IEEEauthorblockA{\IEEEauthorrefmark{4}NVIDIA, Santa Clara, CA, USA}

    \IEEEauthorblockA{\{zelin.xu, y.zhang1, jo.gonzalez, xiaotingsong, whe2, ziboliu, sgchen, zhe.jiang\}@ufl.edu, \\jren03@wm.edu, maitane.olabarrieta@essie.ufl.edu, kasmith@nvidia.com}
}

\maketitle
\thispagestyle{plain}
\pagestyle{plain}
\begin{abstract}
Nearly 900 million people live in low-lying coastal zones around the world and bear the brunt of impacts from more frequent and severe hurricanes and storm surges. Oceanographers simulate ocean current circulation along the coasts to develop early warning systems that save lives and prevent loss and damage to property from coastal hazards.  Traditionally, such simulations are conducted using coastal ocean circulation models such as the Regional Ocean Modeling System (ROMS), which usually runs on an HPC cluster with multiple CPU cores. However, the process is time-consuming and energy expensive. While coarse-grained ROMS simulations offer faster alternatives, they sacrifice detail and accuracy, particularly in complex coastal environments. Recent advances in deep learning and GPU architecture have enabled the development of faster AI (neural network) surrogates. This paper introduces an AI surrogate based on a 4D Swin Transformer to simulate coastal tidal wave propagation in an estuary for both hindcast and forecast (up to 12 days). Our approach not only accelerates simulations but also incorporates a physics-based constraint to detect and correct inaccurate results, ensuring reliability while minimizing manual intervention. 
We develop a fully GPU-accelerated workflow, optimizing the model training and inference pipeline on NVIDIA DGX-2 A100 GPUs. 
Our experiments demonstrate that our AI surrogate reduces the time cost of 12-day forecasting of traditional ROMS simulations from 9,908 seconds (on 512 CPU cores)  to 22 seconds (on one A100 GPU), achieving over 450$\times$ speedup while maintaining high-quality simulation results. 
This work contributes to oceanographic modeling by offering a fast, accurate, and physically consistent alternative to traditional simulation models, particularly for real-time forecasting in rapid disaster response.
\end{abstract}


\section{Introduction}
Nearly 900 million people live in low-lying coastal zones around the world and bear the brunt of impacts from more frequent and severe hurricanes and storm surges~\cite{uncoast}. Oceanographers simulate ocean current circulation along the coasts to develop early warning systems that save lives and prevent loss and damage to property from coastal hazards. Traditionally, such simulations are conducted using coastal ocean circulation models~\cite{gurvan_madec_2023_8167700, mitgcm, HYCOM, zhang2016seamless}. One such model is the Regional Ocean Modeling System (ROMS)~\cite{shchepetkin2005regional}, an ocean model that solves the hydrostatic primitive equations using a Finite Volume scheme. It is widely used by the scientific community for a diverse range of applications, from the coastal and estuarine domains to regional ocean modeling. 

ROMS was first parallelized to run on shared memory systems through multi-threads by the University of California Los Angeles Ocean Research Group~\cite{roms}. To improve the portability across a variety of computer platforms and efficiency for larger model domains and higher resolutions, it was later implemented with a message-passing interface (MPI) by Lawrence Livermore National Laboratory~\cite{wang2005parallel}. The spatial domain is often decomposed into blocks to minimize data dependency and communication costs and balance workloads.  Besides communications among internal grid points, communication is also needed for spatial boundary points with periodic boundaries or physical boundaries. There are several benchmarking studies to evaluate the scalability of the MPI ROMS codes on different numbers of processors~\cite{wang2005parallel}. Although the MPI ROMS codes significantly improve the scalability of coastal simulations, they are still too slow for high-resolution simulation or repeated model runs. Our preliminary results show that with 512 CPU cores on four HPC nodes, simulating 12 days' forecasts on a mesh of 898 by 598 by 12 takes around 9,908 seconds. Considering an ensemble of tens of thousands of models for uncertainty quantification or model calibration to capture extreme events, the total time costs quickly become practically infeasible.

The emergence of Graphics Processing Units (GPUs) has enabled massive parallelism in computing. Several national supercomputers, including Summit, now rely predominantly on GPU-based computation. However, directly porting ROMS MPI codes to the CUDA GPU architecture is non-trivial. The spatial domain can be a non-uniform mesh with varying cell densities. Converting ROMS codes from MPI to GPUs requires additional design and optimization in memory coalescing and thread synchronization. Furthermore, ROMS parallelism is based on MPI, which cannot be directly ported to GPUs. There are several efforts to support ROMS on GPUs~\cite{panzer2013high, mak2011numerical, REMORA}. However, the improvement of these methods is limited, and they fail to exploit the full capabilities of modern GPU hardware.

Because of these barriers, we focus on AI (neural network) surrogates of ROMS simulations on GPUs.  The main idea is to train a deep neural network surrogate on GPUs based on the original ROMS simulation data. Once trained, the surrogate can forecast coastal circulations like the original ROMS model at a much faster speed. In recent years, AI surrogates have been widely used to speed up scientific simulations in a wide range of HPC applications~\cite{krasnopolsky2006complex,o2018using, reichstein2019deep,mohan2018deep,xiao2019reduced,bode2021using,tanaka2021deep, li2023rapid}. For example, vision transformers have been trained for global weather forecasting with promising results~\cite{lam2023learning,fourcastnet,bi2023accurate}. Unfortunately, these surrogate models cannot be directly applied to coastal simulations.

Several unique challenges exist to implement a neural network surrogate for coastal simulations. First, the spatial domain is often a non-uniform grid to capture more subtle physical interactions between the land and the ocean near the coastline. Second, the neural network model needs to capture multi-scale and long-range spatial interactions between different locations as well as the temporal dynamics. Third, running a regional model requires taking in the spatial boundary conditions of the mesh at future snapshots, which is different from global modeling like weather forecasting. Fourth, the large-scale study areas with millions of cells per snapshot lead to substantial data I/O and memory costs, posing significant barriers to efficient model training. Finally, purely data-driven neural networks are often unaware of the existing physical knowledge (e.g., conservation laws), causing spurious predictions. There is a need for a result validation pipeline to evaluate the physical law adherence of AI surrogates.

To address these challenges, this paper introduces an AI surrogate based on a 4D Swin Transformer for coastal circulation modeling in both hindcasts (predicting past variables from historical data) and forecasts (predicting future variables) up to 12 days. Coastal circulation is primarily driven by astronomic tides, wind stress, atmospheric pressure gradients, water density gradients, and freshwater inputs, all of which shape the flow and mixing patterns along coastlines. In this study, to simplify the problem and reduce the complexity of the processes driving the coastal circulation, we focus on characterizing the water level and the flow associated with tidal propagation. We develop a fully GPU-accelerated workflow, optimizing the model training and inference pipeline on NVIDIA DGX A100 GPUs. Our approach not only accelerates simulations but also incorporates a physics-based constraint (the conservation of mass) to detect and correct inaccurate results, ensuring reliability while minimizing manual intervention. 
Our experiments demonstrate that the AI surrogate achieves up to 450$\times$ speedup over traditional ROMS simulations while maintaining high-quality simulation results. The code and data are available at \url{https://github.com/spatialdatasciencegroup/CoastalOceanAISurrogate}. This paper makes the following contributions:
\begin{itemize}
    \item We introduce a 4D Swin Transformer-based AI surrogate, which achieves up to a 450$\times$ speedup on a single NVIDIA DGX A100 GPU in forecasting coastal circulations compared to the traditional MPI ROMS on the same simulation mesh size with 512 CPU cores.
    \item Our AI surrogate achieves an overall promising Mean Absolute Error (MAE) and Root Mean Square Error (RMSE) for all variables on a one-year test dataset compared with ROMS simulation. 
    \item By combining the AI surrogate with MPI ROMS, we design a framework to accelerate end-to-end ROMS simulations while generating reliable forecasting results that adhere to physical laws.
    \item Experimental results show that our AI surrogate exhibits good scalability on multiple GPUs in model training and fast model inference on a single GPU, enabling near real-time coastal forecasting. 
\end{itemize}

\begin{table*}[t]
\centering
\caption{State-of-the-art ROMS Simulation Optimization on HPC Clusters. }
\label{tab:simulation}
 \normalsize
\resizebox{\textwidth}{!}{%
\begin{tabular}{cccccccc}
\toprule
Solution & CPU & RAM & GPU & Optimization Techniques & Simulation Size & Simulation Time & Simulation Overhead \\
\midrule
\cite{wang2005parallel} & 256$\times$ SGI Altix ICE 3700 cores & 512GB & - & Parallelization & 1520 x 1088 x 30 & 3 days & 19,915 seconds \\
\multirow{2}{*}{\cite{jung2021containers}} & \multirow{2}{*}{36$\times$ Intel Xeon 8124-M CPU 
cores} & \multirow{2}{*}{72 GB} & \multirow{2}{*}{-} & \multirow{2}{*}{Containerization} & 422 x 412 x 40 & 3 days & 1,200 seconds \\
 &  &  &  &  & 846 x 826 x 40 & 3 days & 6,000 seconds \\
\cite{Nur_2018} & 32$\times$ Intel Xeon E3-1220 cores & 128GB & - & Containerization & 360 x 400 x 20 & 10 hours & 1,082 seconds \\
\cite{de2021impact} & 128$\times$ Intel Xeon E5-2670 
cores & 384 GB
 & - & IO Parallelism & 212 x 222 x 32 & 1 year & 40 hours \\
 
 Traditional MPI ROMS & 512$\times$ AMD EPYC 7702 cores & 500GB & - & - & 898 x 598 x 12 & 12 days & 9,908 seconds \\
\bf{Our solution} & \bf{16$\times$ AMD EPYC 7742 cores} & \bf{250 GB} & \bf{1$\times$ Nvidia A100} & \bf{AI surrogate} & \bf{898 x 598 x 12} & \bf{12 days} & \bf{22 seconds} \\
\bottomrule
\end{tabular}%
}
\end{table*}

\section{Background}

\subsection{AI Surrogate for physical models}
Existing machine learning techniques for numerical simulations include physics-informed neural networks (PINNs) and neural operator learning. A PINN trains a neural network to solve a partial different equation (PDE) through the automatic differentiation~\cite{raissi2019physics}. PINNs have been tested in simulating coastal flood and wave~\cite{feng2023physics,chen2022physics}, but their training is challenging due to the complex physics-informed regularization terms~\cite{cuomo2022scientific} and the huge computational cost for solving the PDE~\cite{guo2020solving}. Neural operator learning aims to train a neural network surrogate as the solver of a family of PDE instances~\cite{kovachki2023neural}. The surrogate takes the initial or boundary conditions and predicts the solution function. Existing surrogate models include deep convolutional neural networks~\cite{huang2022regional},  graph neural operators~\cite{anandkumar2020neural,li2020multipole}, Fourier neural operators~\cite{li2020fourier,guibas2021adaptive}, DeepONet~\cite{kovachki2023neural}, NodeFormer~\cite{wu2022nodeformer,cao2021choose} and vision transformers~\cite{nguyen2023climax,bi2023accurate}. Some simple methods have been tested for individual coastal problems (e.g., storm surges)~\cite{kim2015time,dong2022recent}. 
There are also methods for irregular spatial points or unstructured meshes in continuous space through implicit neural representation \cite{pan2023neural,yin2022continuous,chen2022crom} or geometric deformation~\cite{li2022fourier}, or fixed-graph transformation~\cite{li2023geometry}, but their neural networks are limited by only taking each sample’s spatial coordinates without explicitly capturing spatial structure. 
There also exists a rich literature that particularly focuses on machine learning for computational fluid dynamics (CFD) and turbulence~\cite{vinuesa2022enhancing}, e.g., accelerating direct numerical simulations \cite{li2022fourier2}, improving turbulence closure modeling \cite{taghizadeh2021turbulence}, and enhancing reduced-order models~\cite{gupta2022three,wang2020reduced}. However, existing research mostly focuses on idealized settings, with few works addressing real-world complex coastal circulation in estuaries. 

Another body of closely related research is AI surrogates for global weather forecasting. For instance, GraphCast \cite{lam2023learning} utilizes Graph Neural Networks (GNNs) for medium-range weather forecasts, presenting a more efficient approach compared to traditional weather simulation systems. It predicts a resolution of 0.25 degrees in longitude and latitude, equivalent to approximately 28km $\times$ 28km at the equator. Similarly, FourCastNet \cite{fourcastnet} uses the Fourier Neural Operator. The Pangu model \cite{bi2023accurate} employs a unique Earth-specific transformer model architecture. Despite the significant advancements of AI surrogates in global weather forecasting, these existing works focus on coarse-scale global forecasting. They cannot be directly applied to regional simulations like ROMS, since ROMS takes not only the initial condition but also the regional boundary conditions of future temporal snapshots. Such boundary conditions are not required in global forecasting models. 

\subsection{ROMS on HPC clusters}
Regional Ocean Modeling System (ROMS) is a semi-explicit ocean model that solves the hydrostatic primitive equations using a Finite Volume scheme. ROMS integrates the Primitive Equations in time, starting at an initial condition and producing, at each time step, updated values of the physical variables (free surface, temperature, salinity, and velocities). It uses a non-uniform, structured spatial discretization (3D Arakawa-C grid) composed of various sigma-coordinate layers, the first of which follows the terrain, the last follows the free surface, and the rest occupies intermediate levels. Time integration is done by decomposing the 3D fields into their barotropic (depth-averaged) and baroclinic (the residual) parts. 
Further details of the forcing in the governing equations of ROMS can be found in~\cite{warner2008development, warner2010development, hsu2017parametric}. 

ROMS is computationally intensive due to the need to discretize numerous state variables across high-resolution spatial grids and store intermediate time levels and halo cells for parallel processing. 
ROMS divides the simulation area into horizontally rectangular zones, assigning each zone to a CPU core. 
As the number of subdivisions increases, the practicality and efficiency of the parallelization diminish. Furthermore, smaller subdivisions lead to larger MPI messages because of the need to share boundary cells among the cores. 

Existing works explore efficient scaling of ROMS by deploying it across multiple nodes on HPC clusters and commercial clouds, leveraging parallelization techniques using MPI~\cite{wang2005parallel}, containerization for reproducibility and portability~\cite{jung2021containers,Nur_2018}, and I/O parallelism to accelerate ROMS execution~\cite{de2021impact}. 
Table~\ref{tab:simulation} summarizes existing efforts in optimizing ROMS simulation on HPC clusters. The table compares the simulation overhead, which represents the computational time required to run the simulation, across different solutions utilizing various computational and memory resources, such as CPUs, GPUs, and RAM. Additionally, the table presents the simulation workload details, including the simulation size (i.e., the mesh size) and the simulation time horizon, which refers to the time duration of coastal circulation simulations.
As can be seen in the table, the existing works~\cite{de2021impact,jung2021containers,wang2005parallel,Nur_2018} exhibit significantly higher computational overheads. For example, in the first row~\cite{wang2005parallel}, running ROMS for 3-day forecasting on a $1520\times1088\times 30$ with 256 CPU cores took 19,915 seconds. This result is consistent with our benchmarking results of running ROMS simulation on a $898\times 598\times 12$ mesh with 512 cores, which took $9,908$ seconds (half of the time cost of~\cite{wang2005parallel}). In contrast, 
our AI surrogate solution demonstrates remarkable efficiency in optimizing ROMS simulations. By leveraging a combination of 16 AMD EPYC 7742 Processors and one NVIDIA A100 GPU, AI surrogate achieves a simulation overhead of just 22 seconds for 12-day forecasts of coastal ocean circulation with a mesh size of 898$\times$598$\times$12, which is near \textbf{450$\times$ faster} than traditional MPI ROMS with the same simulation size. 
This indicates a significant potential for performance improvement by adapting AI surrogates in ROMS simulations.

\section{Design}
\begin{figure*}[t]
\centering
\includegraphics[width=0.9\textwidth]{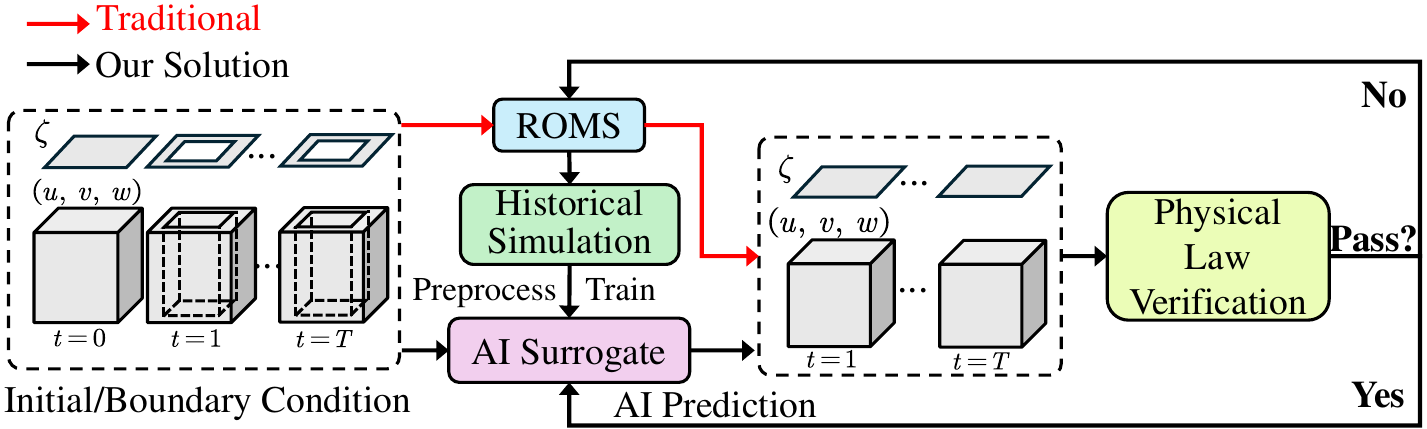}
\caption{The overall workflow of our method. ROMS historical simulations are pre-processed and used to train the AI surrogate. The trained AI surrogate takes initial and boundary conditions as inputs and predicts the interior values. A verification module checks physical law adherence of AI predictions and switches to ROMS if the surrogate fails the check. }
\label{fig:workflow}
\end{figure*}

\subsection{Overall Workflow}
\label{sec:workflow}

The overall workflow is shown in Figure~\ref{fig:workflow}. Historical coastal simulations from ROMS are pre-processed and used to train an AI surrogate. Once trained, the AI surrogate can make forecasts by taking the initial condition of physical variables at time $t_0$ and the boundary conditions of those variables from time $t_1$ to time $t_T$ as inputs. For example, to forecast 24 snapshots (one snapshot every half hour)  of variables inside a 3D mesh from 12:00 p.m. on March 26, 2012, the initial condition is the state of variables on the entire 3D mesh at a previous snapshot (11:30 a.m. on March 26, 2012). Boundary conditions include variables at the lateral boundary of the 3D mesh in the 24 snapshots to be forecasted, spanning up to 12:00 a.m. on March 27, 2012. The surrogate predicts the interior values of those variables at time $t_1$ to $t_T$ as outputs. Some variables, such as free surface elevation, are 2D, while others, such as circulation velocities, are 3D. 
A physics-based verification module checks if the surrogate's forecast satisfies the physical constraints, such as the conservation of mass. If the forecasts meet these requirements, the forecasting continues with the surrogate generating the next episode of snapshots. Otherwise, the pipeline temporarily switches back to the original ROMS model to run the forecast. 

For long-term forecasting, we employ a dual-model approach, i.e., training two models with the same architecture but at different temporal resolutions. Specifically, one model makes forecasts over 12 days with 24 snapshots (one snapshot every 12 hours), while the other model makes forecasts over 12 hours with 24 snapshots (one snapshot every half an hour). To generate 12-day forecasts with half-hour intervals (576 snapshots), we combine the two models, i.e., first running the coarse model to generate a 12-day forecast with a snapshot every half day and then using each coarse snapshot as the initial condition to generate 24 half-hour snapshots via the second model. We set 12 days as the prediction limit according to oceanographers for practical use. The 12-day prediction limit is set based on practical considerations from oceanographers, as boundary condition uncertainties increase over time, making longer forecasts increasingly unreliable, even for ROMS.

\subsection{Data preparation}
The dataset for this study is derived from our decade-long (2008-2017)  Regional Ocean Modeling System (ROMS) simulation in Charlotte Harbor. This comprehensive dataset includes temporal snapshots every \textit{half-hour}, featuring over ten physical variables such as circulation velocities (horizontal and vertical), temperature, salinity, and free surface elevation, alongside several external force variables, including surface wind components, air pressure, relative humidity, solar radiation, and rainfall rate. The spatial domain is represented by a non-uniform 3D mesh of $898 \times 598 \times 12$ with a higher resolution near river channels and inlets to capture the complex land-water interactions. For simplicity, we focus on the tidal wave propagation process in the estuary, which involves a subset of four physical variables, including circulation velocities along two horizontal directions ($u, v$) and the vertical direction ($w$), as well as the free surface elevation ($\zeta$).

We use simulations in 2011 for training and validation (9:1) and simulations in 2012 for testing. To train the 12-hour forecasting model, we augment the 2011 dataset using a sliding window of 24 time steps (12 hours) with a stride of 6. This results in 2,588 training instances, 288 validation instances, and 720 test instances. Similarly, to train the 12-day model, we resample the 2011 data with 12-hour intervals, generating 2,509 training instances and 279 validation instances. The volumes of the two training datasets are 2.6 TBs and 2.5 TBs, respectively.  The 2012 data was partitioned into non-overlapping windows for testing in both cases. 

In the original ROMS simulation data, the current velocity variables are located on the sides of cells, while the other variables are located in cell centers. To accommodate neural network training, we use linear interpolation to resample all variables to cell centers. Additionally, zero-padding is used to adjust the mesh size to $900 \times 600 \times 12$. The original ROMS simulations are based on FP64 for numerical stability, but the data is converted to FP16 for model training to enable faster computation and reduced memory usage. 
Numerical models like ROMS require higher precision due to accumulated rounding errors in numerous iterations to solve complex equations, while AI models can capture patterns from data without iterative operations. Further studies are needed to see the effect of such data compression on the accuracy performance of AI surrogates.
All variables are normalized using z-score normalization based on the mean and standard deviation from the 2011 data.

\subsection{AI surrogate architecture} 
\begin{figure}[t]
\centering
\includegraphics[width=0.45\textwidth]{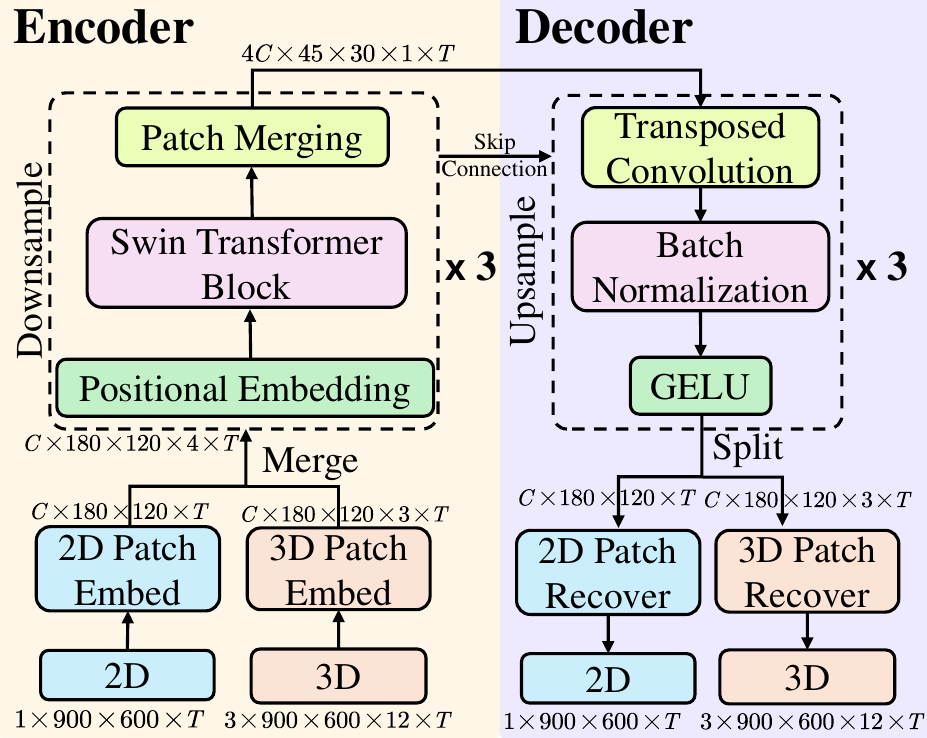}
\caption{The architecture of the AI surrogate. The encoder processes 2D and 3D variables via patch partitioning and patch embedding, then merges them along the depth dimension. The 4D Swin Transformer with Positional Embedding and Patch Merging downsamples the 4D data. The decoder employs Transposed Convolutions, Batch Normalization, GELU activation, and Skip Connection from the corresponding encoder layers to upsample the variables and finally recover them to their original spatial and temporal size.}
\label{fig:architecture}
\end{figure}

Our AI surrogate model is built upon the 4D Swin Transformer model~\cite{liu2021Swin, liu2022video, kim2024swift}. It has an encoder-decoder architecture, as shown in Figure~\ref{fig:architecture}. The model starts by taking the initial conditions at time $t=0$ and boundary conditions from $t=1$ to $T$ across all simulation variables (both 3D variables $u,v,w$ and 2D variable $\zeta$ on the mesh surface). 
To efficiently process this high-dimensional data with millions of mesh cells, we first employ patch partitioning and embedding~\cite{dosovitskiy2020image} for 3D and 2D variables. Patch partitioning divides the 3D/2D spatial domain into smaller blocks, effectively reducing the spatial resolution. Patch embedding transforms these blocks into the same latent representation space, facilitating the concatenation of a 2D variable (free surface elevation $\zeta$) and 3D variables (current velocity $u,v,w$) along the depth dimension. 
After considering an additional temporal dimension, the 4D data was then fed into a 4D Swin Transformer encoder. It leverages the 4D spatiotemporal attention mechanism to learn complex dependency patterns across both space and time with advanced window self-attention and shifted-window self-attention layers for computational efficiency, simultaneously downsampling the data with patch merging operations to capture the multi-scale information.
The decoder employs multiple layers of transposed convolutions, with batch normalization~\cite{ioffe2015batch} and Gaussian Error Linear Units (GELU)~\cite{hendrycks2016gaussian} activations to gradually upsample the encoded data, restoring it to its original resolution. A skip connection operation similar to the U-Net architecture~\cite{ronneberger2015u, hatamizadeh2021swin} is introduced to fuse fine-grained information directly from the corresponding encoder layers. Finally, the model splits the 3D and 2D variables after upsampling and recovers them to the original spatial and temporal shape.

{\bf Patch embedding.} To efficiently handle high-dimensional data with millions of mesh cells, we first employ patch partitioning and embedding, dividing the 3D/2D spatial domain into smaller segments and embedding these segments into the same embedding space. Choosing the appropriate patch size is crucial: smaller patches can capture fine-grained information more effectively but increase the number of patches, thereby raising the computational cost of the following attention mechanisms. Thus, a balance must be made between model accuracy and training complexity. In this study, 3D variables ($u, v, w$) form a $3 \times H \times W \times D \times T$ tensor, while the 2D variable ($\zeta$) contains a $1 \times H \times W \times T$ volume, where $H, W, D$, and $T$ are height, width, depth, and time respectively. These variables are partitioned into patches and mapped into a $C$-dimensional latent space. For the 3D variables, the patch size is $P_H \times P_W \times P_D$, resulting an embedded shape of $C \times \frac{H}{P_H}\times \frac{W}{P_W} \times \frac{D}{P_D} \times T$. For the 2D surface variable, the patch size is $P_H \times P_W$, so the embedded data have a shape of $C \times \frac{H}{P_H}\times \frac{W}{P_W}  \times T$, where $C$ is the initial embedding dimension. These embedded volumes are then concatenated along the depth dimension and passed to the encoder.

{\bf Swin transformer block.}
\begin{figure}[t]
\centering
\subfigure[An example of window shifting.]{
\includegraphics[width=0.28\textwidth]{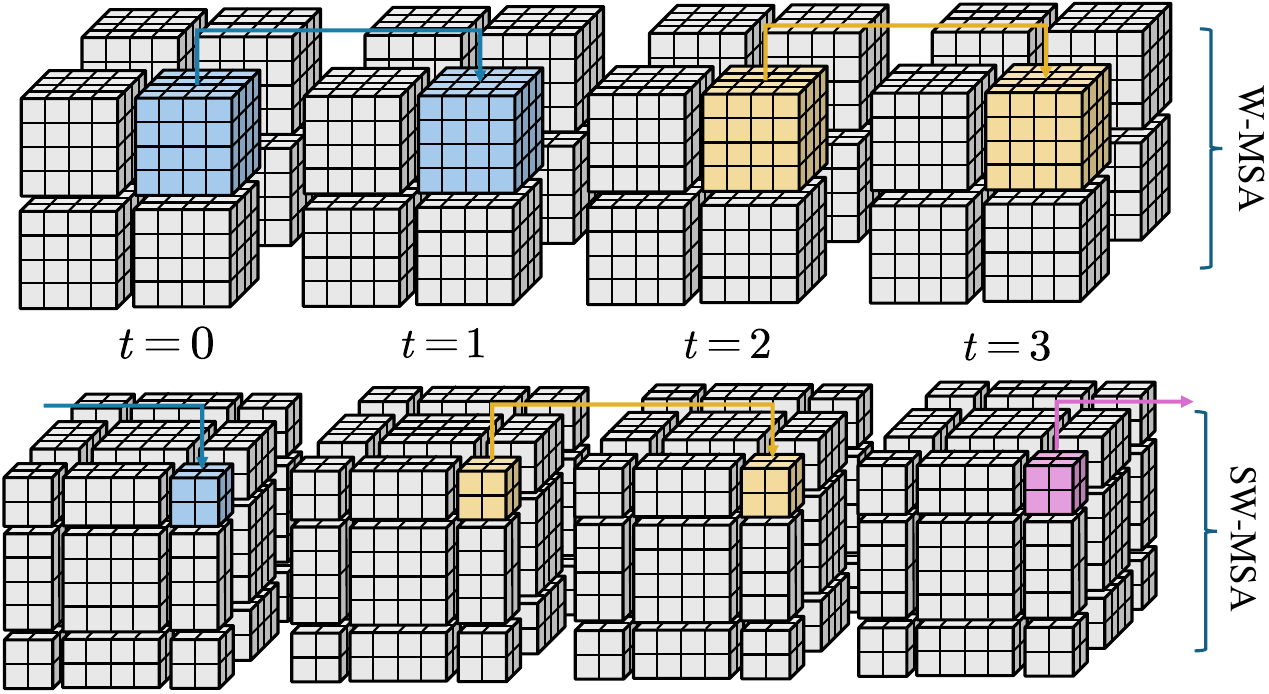}
\label{fig:shift_window}
}
\subfigure[Swin transformer blocks.]{
\includegraphics[width=0.17\textwidth]{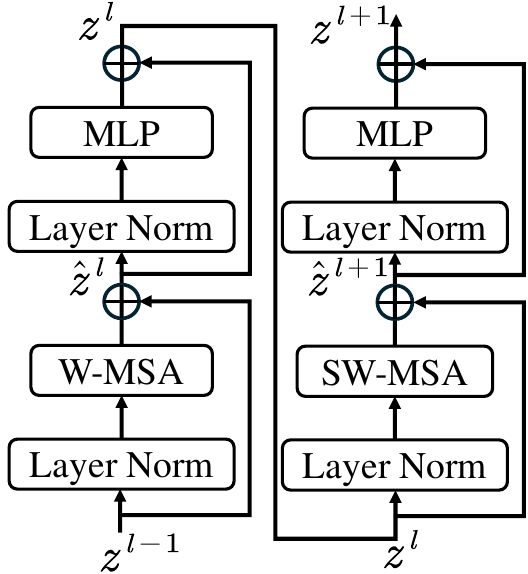}
\label{fig:swin_layer}
}
\caption{Swin transformer block. (a) In this example, the size of input patches and the windows are $8 \times 8\times 8 \times 4$ and $4 \times  4 \times  4 \times 2$, respectively. Following the window partitioning and shifting methods, the numbers of grouped windows in W-MSA and SW-MSA become $2\times 2\times 2\times 2 =16$ and $3 \times  3 \times  3 \times  3 = 81$, respectively. (Source: \cite{kim2024swift, liu2022video})}
\label{fig:swin}
\end{figure}
To effectively process high-dimensional spatiotemporal data in coastal simulations, we adopt the 4D Swin Transformer architecture. This approach is well-suited for the task because it can capture complex spatial and temporal dependencies while mitigating the computational challenges associated with traditional attention mechanisms.

Multi-head self-attention (MSA)~\cite{vaswani2017attention} has revolutionized modeling capabilities by allowing the network to focus on different parts of the input simultaneously. It achieves this by computing attention scores using queries ($Q$), keys ($K$), and values ($V$), derived from the input data
\begin{equation}
    \text{Attention}(Q, K, V) = \text{softmax}(\frac{QK^T}{\sqrt{d}})V
\end{equation}
where $d$ is the embedding dimension. MSA extends this concept by splitting the attention mechanism across multiple heads, each computing attention independently:
\begin{equation}
    \text{MSA}(Q, K, V) =  \text{Concat}( \text{head}_1, \dots, \text{head}_h)W^{O}
\end{equation}
where head$_i$ = Attention$(QW_i^{Q}, KW_i^{K}, VW_i^{V})$, $W^{O}$, $W_i^{Q}$, $W_i^{K}$, $W_i^{V}$ are learnable parameter matrices, and $h$ is the number of heads.

While MSA is powerful, it has an $O(n^2)$ computational complexity to the number of patches, making it unsuitable for high-dimensional simulation data due to prohibitive computational cost. The Swin Transformer addresses this challenge by partitioning the input into smaller, non-overlapping windows and applying attention within these local windows.  

The core of the Swin Transformer model is the \textit{window multi-head self-attention (W-MSA)} layer. Given $H'\times W' \times D' \times T$ input patches, the patches are partitioned by a window of size $M_H \times M_W \times M_D \times M_T$, resulting in $\frac{H'}{M_H}\times \frac{W'}{M_W} \times \frac{D'}{M_D} \times \frac{T}{M_T}$ non-overlapping windows. While W-MSA improves efficiency, it does not allow for information exchange between windows. To address this, we use \textit{shifted window multi-head self-attention (SW-MSA)} layer which shifts the windows by a set number of patches in the subsequent layer to enable cross-window connections for a more comprehensive understanding of the data. Although the number of windows increases in SW-MSA, the actual computation cost remains similar by leveraging the cyclic-shifting batch~\cite{liu2021Swin}. An example of detailed operations of 4D W-MSA and SW-MSA is shown in Figure~\ref{fig:shift_window}. Combining the W-MSA layer and the SW-MSA layer, two successive 4D Swin Transformer blocks, as shown in Figure~\ref{fig:swin_layer}, are computed as the following:
\begin{equation}
    \begin{split}
    & \hat{z}^l = \text{W-MSA}(\text{LN}(z^{l-1})) + z^{l-1},\\
    & z^l = \text{MLP}(\text{LN}(\hat{z}^{l})) + \hat{z}^{l}, \\
    & \hat{z}^{l+1} = \text{SW-MSA}(\text{LN}(z^{l})) + z^{l},\\
    & z^{l+1} = \text{MLP}(\text{LN}(\hat{z}^{l+1})) + \hat{z}^{l+1},
    \end{split}
\end{equation}
in which (S)W-MSA, LN, and MLP denote the (Shifted) Window Multi-head Self-Attention, Layer Normalization~\cite{ba2016layer}, and Multi-Layer Perceptron module, respectively. Moreover, $\hat{z}^l$ and $z^l$ denote the output features of the (S)W-MSA module and the following MLP module for block $l$, respectively.

{\bf Patch merging.}
Patch merging is used to downsample the data, enabling a hierarchical feature extraction structure while reducing the computational complexity of subsequent layers.  
\begin{figure}
\centering
\vspace{-3mm}
\includegraphics[width=0.25\textwidth]{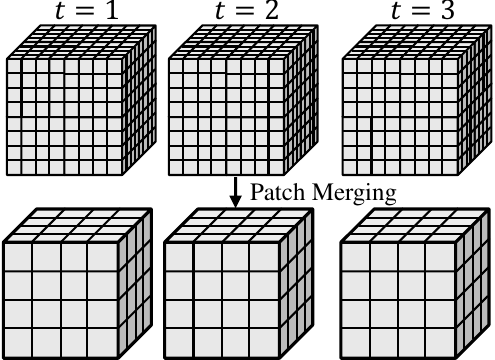}
\caption{Patch merging is applied to the spatial dimensions, while the temporal dimension remains unchanged. In this example, a tensor of size 8$\times$8$\times$8$\times$3 is merged to 4$\times$4$\times$4$\times$3.}
\label{fig:merge}
\vspace{-3.5mm}
\end{figure}
Following prior works~\cite{liu2022video, tang2022self, kim2024swift}, the patch merging step performs on the three spatial dimensions (height, width, depth) but not for the temporal dimension. During the patch merging operation, the spatial dimensions are reduced by half, and the channel size is doubled as compensation. 
As shown in Figure~\ref{fig:merge}, consider a tensor with dimensions $C'\times H'\times W' \times D' \times T$. During patch merging, this tensor is reshaped into dimensions $8C'\times \frac{H'}{2}\times \frac{W'}{2} \times \frac{D'}{2} \times T$, where $2 \times 2 \times 2$ spatially neighboring patches are concatenated along the channel dimension. Then, each channel in the resulting tensor is projected onto a $2C'$ dimensional space by applying a single fully connected layer, resulting $2C'\times \frac{H'}{2}\times \frac{W'}{2} \times \frac{D'}{2} \times T$ in total. 

{\bf Positional encoding.}
To accurately represent the spatiotemporal distribution of data on non-uniform grids, we use absolute positional encoding, which explicitly captures the positions of patches in both space and time.  In line with~\cite{bertasius2021space}, we separately add positional embeddings for the spatial and temporal dimensions. Specifically, given an input tensor with dimensions of $C'\times H'\times W' \times D' \times T$, we define spatial and temporal positional embedding tensors with dimensions of $C'\times H'\times W' \times D' \times 1$ and $C'\times 1\times 1 \times 1 \times T$, respectively. These tensors are then added to the input tensor using broadcasting. 

{\bf Decoder.}
After the encoding process, the decoder reconstructs the spatial and temporal features from their embedded space, enabling the generation of high-resolution outputs. It starts by applying a series of 3D transposed convolutional layers with batch normalization and Gaussian Error Linear Unit (GELU) activations to upsample the embedded tensor back to its original dimensions. At this stage, a skip connection similar to the U-Net architecture~\cite{ronneberger2015u, hatamizadeh2021swin} is introduced, allowing the decoder to fuse spatial information directly from the corresponding encoder layers, which helps preserve fine-grained spatial details and improves the overall reconstruction quality. Following this step, the decoder splits the 3D and 2D patches and applies different patch recovery methods, which contain a 3D/2D transposed convolution, batch normalization, and GELU activation, with $1\times1$ 3D/2D convolutions to reconstruct the spatiotemporal features and align the features with their original shape.

\begin{table*}[t]
   \caption{Memory requirement in AI surrogates training} 
   \label{tab:mem_cost}
   \small
   \centering
   \begin{tabular}{lccc}
   \toprule\toprule
   \textbf{} & \textbf{Training Sample Loading} & \textbf{Training Sample Processing} & \textbf{Model Parameter Updating} \\ 
   \midrule
   Memory Requirements & 4GB & 42GB  & 12GB \\
   Data Stores & SSD $\rightarrow$ CPU memory $\rightarrow$ GPU memory & GPU memory & GPU memory\\
  Accessing Throughput & 750 MB/s & 2TB/s & 2TB/s\\
   \bottomrule
   \end{tabular}
\end{table*}

\subsection{Offline AI surrogate training}
\label{sec:training_optim}

AI surrogate model training on high-resolution spatiotemporal simulations requires intensive I/O to load a large volume of training data into GPU memory. To fully utilize the parallelism of GPUs and achieve efficient training with large datasets, it is essential to identify and address performance bottlenecks. We analyze the throughput of each stage in the training pipeline, including training sample loading, training sample processing, and model parameter updating, to determine the limiting factor. Table~\ref{tab:mem_cost} shows results.

Table~\ref{tab:mem_cost} lists the memory requirement for each stage in the training pipeline when we process one sample with mesh size as 900$\times$600$\times$12 using mixed precision training. 
We identify two bottlenecks for AI surrogate training. (1) Low bandwidth of SSD slows down training sample loading from SSD to higher memory hierarchy. Loading one sample from SSD to CPU memory causes a 5.5-second latency on the critical execution path. 
(2) Training batch size is limited by the capacity of GPU since the activation tensor generated in each forward pass consumes most GPU memory. 
Given the NVIDIA A100 GPU memory capacity limitation of 80GB, we can only use per GPU training batch size as 1 for AI surrogate training, which significantly impacts the convergence efficiency.

We propose two optimization techniques to address the above performance bottlenecks by (1) reducing I/O strain by leveraging OS-level caching; and (2) reducing peak memory consumption of training with customized activation checkpoint.
By addressing the I/O and memory usage bottlenecks head-on, we pave the way for more scalable, efficient, and effective AI surrogate training.

{\bf Leveraging OS cache for I/O strain reduction.} An additional optimization was introduced via the PyTorch DataLoader's prefetch technique. By increasing the number of subprocesses dedicated to data loading, we significantly improved the efficiency of moving data from SSD storage to CPU RAM. This adjustment ensures that data is pre-fetched and ready for processing in parallel with ongoing computations, effectively hiding I/O overhead during model training. Such an approach is crucial in maximizing computational throughput.

Further, to optimize the transfer of data from CPU to GPU, we employed the use of pinned memory and non-blocking I/O operations. Pinned memory, or page-locked memory, facilitates faster data transfer rates by preventing the paging of memory, while non-blocking I/O operations allow for concurrent execution of data transfers and computation. These modifications substantially accelerate the data pipeline, mitigating potential bottlenecks between CPU and GPU.

{\bf Reduce GPU peak memory consumption through activation checkpointing.} A pivotal aspect of our optimization strategy involves the implementation of checkpointing techniques to effectively manage GPU memory usage. By selectively storing the activations of the Swin Transformer layers, we can reconstruct the intermediate states as needed, rather than maintaining them in memory throughout the training process. To minimize the reconstruction overhead while minimizing the memory usage for activations, we store the activations of the shifted-window multi-head self-attention (SW-MSA) layers but discard the activations of other layers. This is because the SW-MSA layer is the most computationally expensive, but the activations of the SW-MSA layer and the remaining layers have the same size.
This technique not only reduces the per-sample GPU memory footprint but also enables us to increase the per-GPU batch size, effectively doubling the computational workload without necessitating additional GPU memory resources.

\subsection{Inference and verification}

One crucial aspect of the AI surrogate's inference is result verification. A verification module grounded in fundamental physical laws represents a major innovation in our AI surrogate system. By ensuring that the surrogate's predictions adhere to \textit{mass conservation law}, we establish a rigorous framework for evaluating and improving the quality of our forecasting results. This verification process, coupled with a dynamic feedback loop involving both our AI surrogate and ROMS, underscores our commitment to developing a predictive system that is both scientifically robust and computationally efficient.

{\bf Verification to ensure adherence to physical laws.}
At the core of our verification strategy is the mass conservation law of water, as it is straightforward and reliably tests one important aspect of the AI model's physical consistency. To satisfy this law, the divergence of circulation velocities into a water column (water inflow) should be equal to the time derivative of free surface elevation (increase of volume). For a horizontal domain $\Omega$ of contour $\Gamma$, the mass conservation law can be expressed as Equation~\ref{eq:mass},
\begin{equation}\label{eq:mass}
    \frac{\partial}{\partial t} \int_{\Omega} (h + \zeta) \, d\Omega = \int_{\Gamma} (h + \zeta) \, \mathbf{u} \cdot \mathbf{n} \, d\Gamma
\end{equation}
where $h$ is the depth (it is a constant for the same study area), $\mathbf{u}$ is the horizontal velocity vector (a vector of $u, v$), $\mathbf{n}$ is the outward-pointing unit normal vector to the boundary, and the horizontal domain $\Omega$ is a horizontal grid cell in our case. This physical law serves as a critical metric for assessing the quality of predictions generated by our AI surrogate. By integrating this verification step, we can evaluate whether the surrogate's outputs are physically consistent. Specifically, we compute the residual value by subtracting both sides of the equation and taking the absolute value:
\begin{equation}
    \text{WaterMassResidual} = \bigg |\frac{\partial}{\partial t} \int_{\Omega} (h + \zeta) \, d\Omega - \int_{\Gamma} (h + \zeta) \, \mathbf{u} \cdot \mathbf{n} \, d\Gamma\bigg |
\end{equation}
We compute this value for all cells in one forecast. If the average value is less than a setting threshold, then the prediction is considered physics-consistent and can pass the verification.


{\bf Combine AI and ROMS for correction and improvement.}
The verification module checks the AI surrogate's predictions against established conservation criteria. 
For instances in which the AI surrogate's results fail to meet the verification criteria, indicating a deviation from expected physical behaviors, our system is designed to revert to the Regional Ocean Modeling System (ROMS) for direct simulation. This fallback mechanism ensures that our predictions remain anchored to robust, physics-based simulations, maintaining the reliability of our forecasts. A key factor that will impact the scaling performance of our pipeline is the passing rate of the AI surrogate in physics verification. A high passing rate will lead to a smaller chance of reverting back to the original ROMS simulation and thus achieving a better speed-up.


\section{Evaluation}
\begin{table*}[t]
\centering
\caption{MAE and RMSE of the AI surrogate in 12-hour and 12-day forecasts.}
\label{tab:overall}
\begin{tabular}{lcccccccc}
   \toprule\toprule
\multirow{2}{*}{\textbf{Time Horizons}} & \multicolumn{4}{c}{\textbf{Mean Absolute Error (MAE)}} & \multicolumn{4}{c}{\textbf{Root Mean Squared Error (RMSE)}} \\ 
   & \textbf{$u$} [m/s] & \textbf{$v$}  [m/s] & \textbf{$w$}  [m/s] & \textbf{$\zeta$} [m]  & \textbf{$u$}  [m/s] & \textbf{$v$}  [m/s] & \textbf{$w$}  [m/s] & \textbf{$\zeta$}  [m] \\ 
    \midrule
\textbf{12 hours}  & 1.80E-02 & 1.73E-02 & 9.60E-05 & 4.58E-02 & 2.89E-02 & 2.61E-02 & 3.57E-04 & 7.25E-02 \\ 
\textbf{12 days} & 1.49E-02&	1.40E-02&	8.27E-05&	4.79E-02&	2.50E-02&	2.10E-02&	2.61E-04&	7.74E-02\\ 
\bottomrule
\end{tabular}%
\end{table*}

\begin{figure*}[t]
\centering
\subfigure[Horizontal current velocity $u$ (east-west)]{
\includegraphics[width=0.65\textwidth]{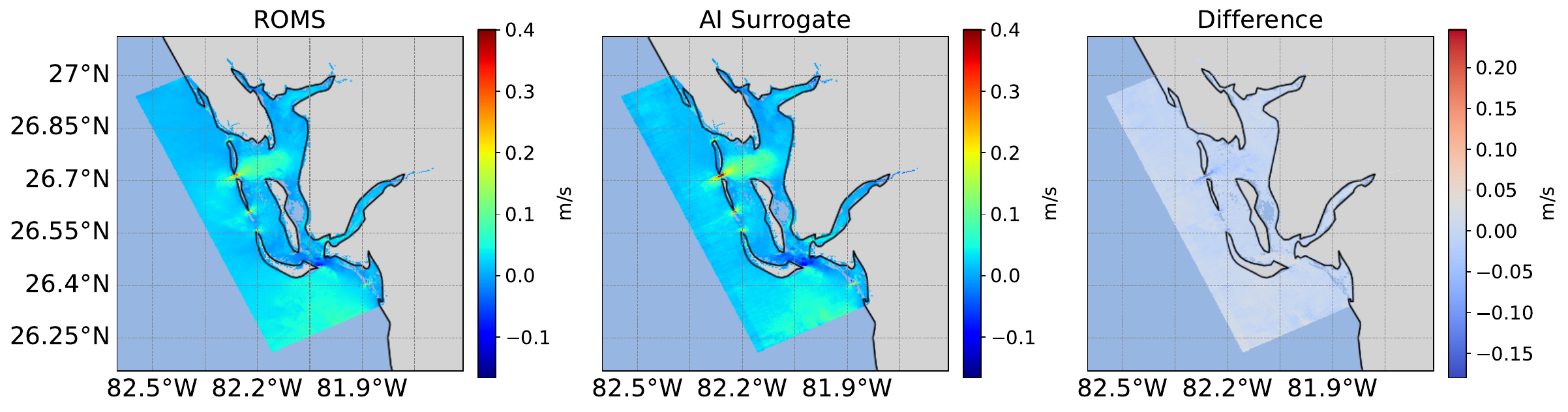}
}
\centering
\subfigure[Horizontal current velocity $v$ (north-south)]{
\includegraphics[width=0.65\textwidth]{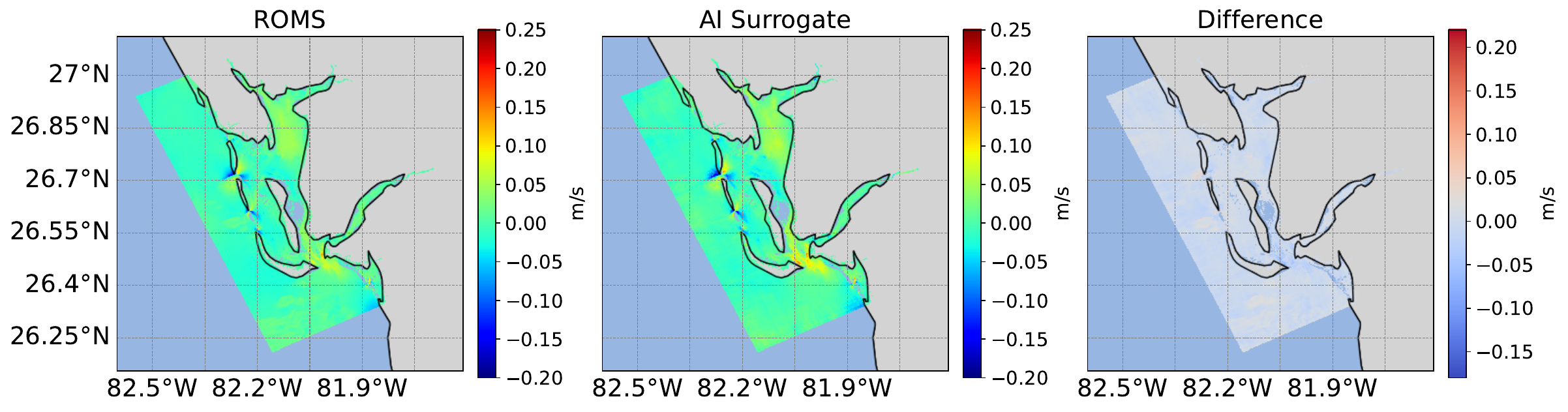}
}
\centering
\subfigure[Free surface elevation $\zeta$]{
\includegraphics[width=0.65\textwidth]{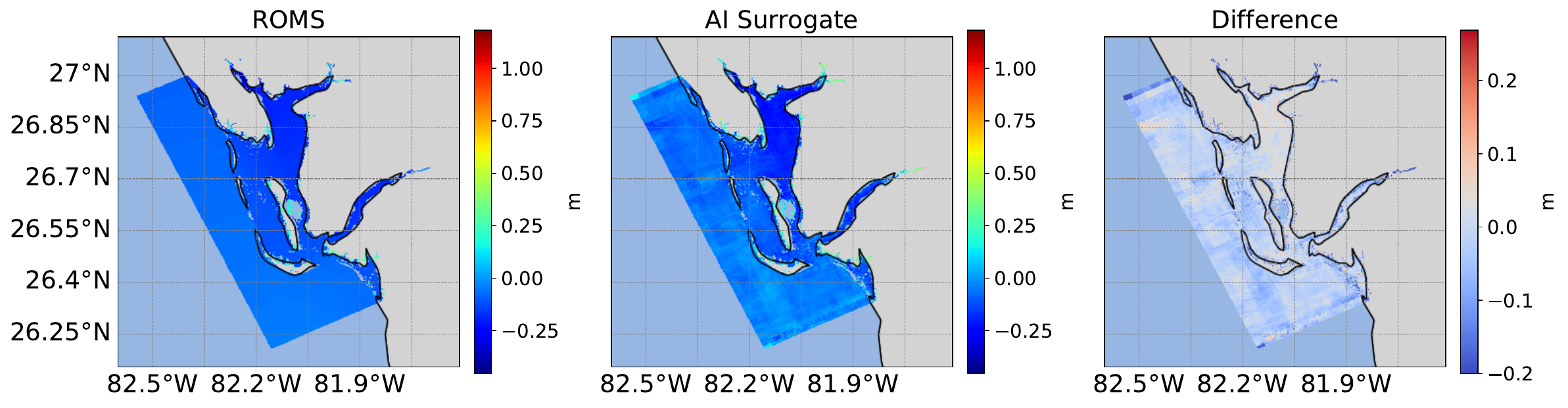}
}
\caption{Visualization of ROMS and AI surrogate forecasting results for $u, v$, and $\zeta$ at 12:30 p.m. on April 1, 2012  ($u,v$ are on the surface level), using the initial condition from 12 p.m. on March 26, 2012. The AI surrogate accurately captures some of the spatial patterns in coastal circulation.}
\label{fig:visual_map}
\end{figure*}

\subsection{System hardware platform}
Our computational experiments are executed on an HPC cluster with 140 NVIDIA DGX-2 nodes. Each node is equipped with eight NVIDIA A100 GPUs (each A100 has 80GB of GPU memory), alongside 2 AMD EPYC 7742 64-Core Processors (128 CPU cores for each node) and 2010GB of CPU memory. The GPUs within each node are intricately linked through Nvlink technology, facilitating efficient intra-node communication. The cluster employs a sophisticated network of 10x HDR InfiniBand (IB) interfaces for inter-node data transfer, designed to ensure seamless, non-blocking communication across nodes.
In the training phase of our AI surrogate model, we use up to 32 A100 GPUs within four DGX-2 nodes. 
For inference, our model's requirements are significantly reduced, necessitating only a single A100 GPU complemented by 16 CPU cores and 250 GB of memory, highlighting the model's efficiency.

\begin{figure*}[t]
\centering
\begin{minipage}{0.32\textwidth}
    \centering
    \subfigure[Selected Locations]{
        \includegraphics[width=\textwidth]{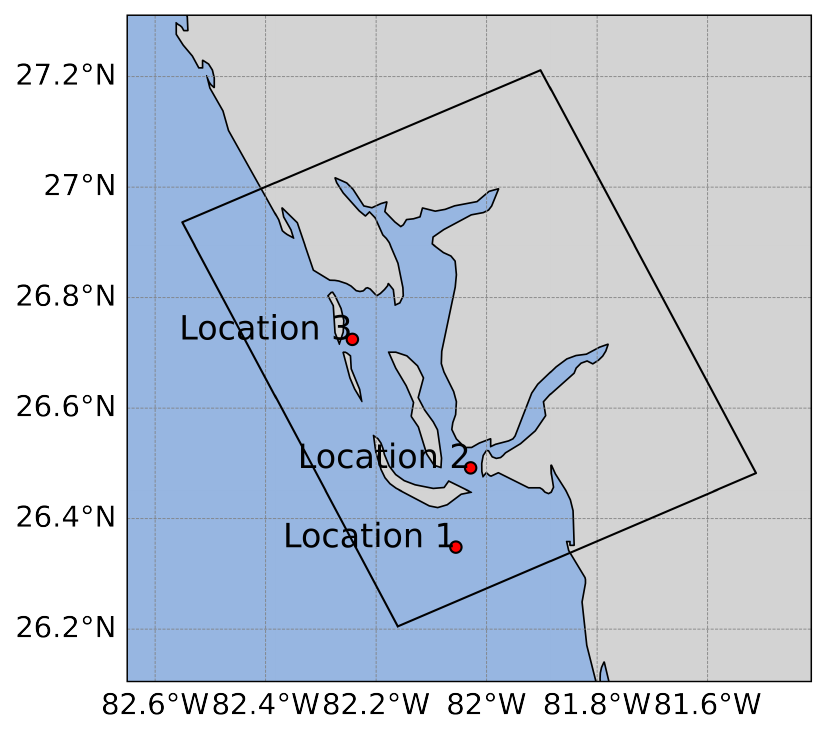}
    }
\end{minipage}
\begin{minipage}{0.63\textwidth}
    \centering
    \subfigure[Location 1 ($26.35\degree N, 82.06\degree W$)]{
        \includegraphics[width=\textwidth]{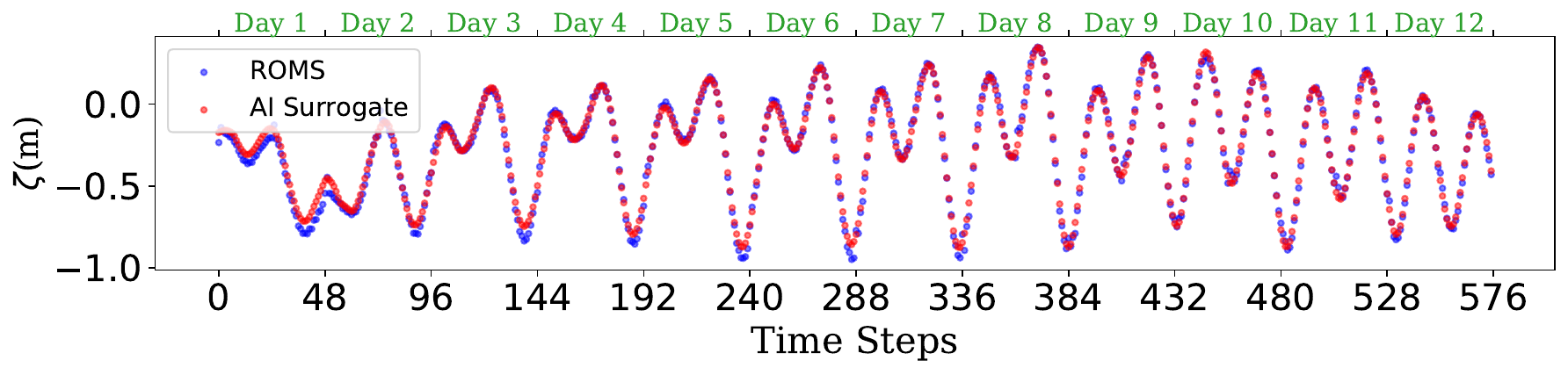}
    }
    \subfigure[Location 2 ($26.49\degree N, 82.03\degree W$)]{
        \includegraphics[width=\textwidth]{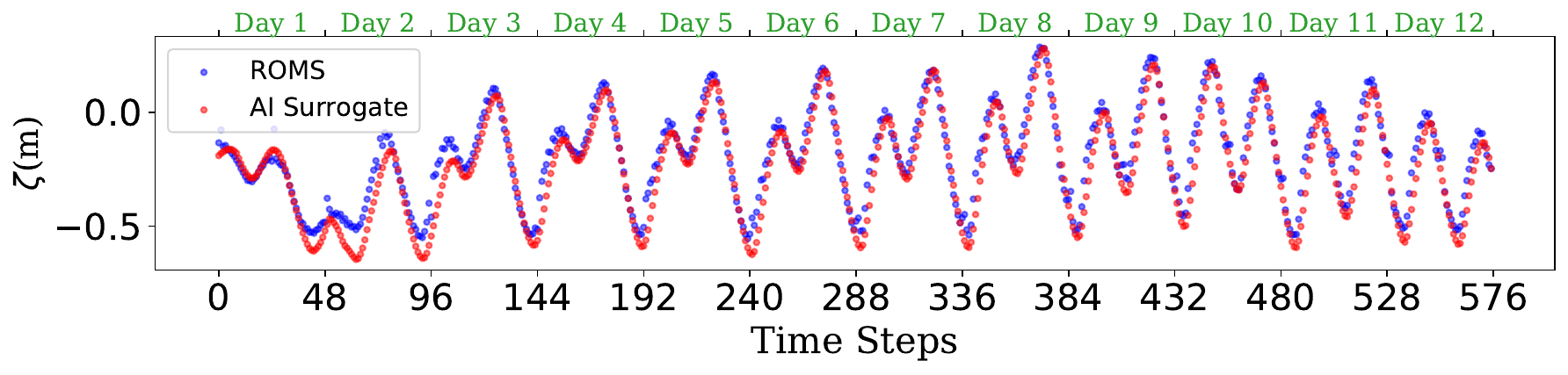}
    }
    \subfigure[Location 3 ($26.72\degree N, 82.24\degree W$)]{
        \includegraphics[width=\textwidth]{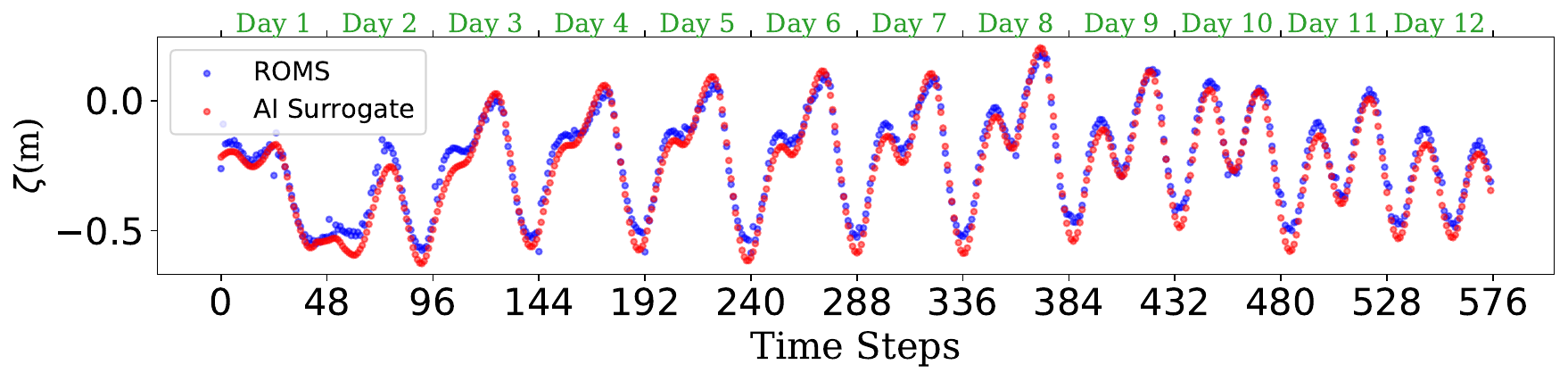}
    }
\end{minipage}
\caption{Comparison of ROMS simulation and AI surrogate predictions for $\zeta$ over time at three randomly selected locations from 12 p.m. January 2, 2012, to 12 p.m. January 14, 2012. The time interval between steps is 30 minutes, resulting in 48 time steps per day. The AI surrogate's predictions closely match the ROMS simulations over the entire 12-day forecast period, indicating that the surrogate successfully captures the temporal dynamics of the coastal circulation.}
\label{fig:scatter}
\end{figure*}

\subsection{Overall Performance}  
In our experiments, the AI surrogate model is designed to generate predictions based on initial conditions and boundary conditions for the next 24 time steps. With two different time intervals (0.5-hour and 12-hour), the model can produce results spanning 12 hours and 12 days, both at a 0.5-hour resolution. Table~\ref{tab:overall} presents the average MAE and RMSE for prediction performance on the 2012 test data. 

The AI surrogate model handles both 3D and 2D variables. For 3D variables, we use a patch size of $5 \times 5 \times 4$ for 3D variables and $5 \times 5$ for 2D variables, with an initial patch embedding size of 24. The backbone consists of 4D Swin Transformer operations, applied iteratively three times, and employed attention heads of 3, 6, and 12, respectively. The initial window attention operation utilizes a window size of (4, 4, 2, 2) (for 4 dimensions) while subsequently adopting a window size of (2, 2, 2, 2).

In alignment with the optimization strategies outlined in Section~\ref{sec:training_optim}, we implement pin memory and non-blocking I/O to streamline data transfer between CPU and GPU, reducing bottlenecks and improving computational efficiency. During training, we use a batch size of 2 per GPU and employ 6 worker processes with a prefetch factor of 2 to ensure continuous data loading. Both fine- and coarse-resolution models are trained for 30 epochs, with training times of 58 and 56 minutes, respectively. 

For inference on a single A100 GPU, a 12-hour forecast takes 0.888 seconds. For the 12-day forecast, which involves running the coarse-resolution model followed by 24 iterations of the fine-resolution model, the total inference time is 22.2 seconds.

\subsection{Case Study}

To demonstrate the forecast capabilities of our AI surrogate system, we present the visualization of spatial distributions of the $u, v$ and  $\zeta$ variables in Figure~\ref{fig:visual_map}. The vertical velocity $w$ is always very small and is very close to 0 for most areas, so we do not show the visualization of $w$ here. We also show the temporal variation of ROMS's and AI surrogate's output on free surface elevation ($\zeta$) at three distinct locations from 12 p.m. January 2, 2012, to 12 p.m. January 14, 2012, as shown in Figure~\ref{fig:scatter}. These visualizations underscore the system's proficiency in capturing the complex spatial and temporal dynamics inherent in the ROMS simulation.

\begin{table*}[ht]
\centering
\caption{Sensitivity analysis of the patch size.}
\label{tab:sensitivity}
\resizebox{0.95\textwidth}{!}{
\begin{tabular}{ccccccccccc}
   \toprule\toprule
\multirow{2}{*}{\textbf{Patch Size}} & \multirow{2}{*}{\textbf{\# of Parameters [M]}} & \multirow{2}{*}{\textbf{Time/Instance [s]}} & \multicolumn{4}{c}{\textbf{MAE}} & \multicolumn{4}{c}{\textbf{RMSE}} \\ 
 &  &  & \textbf{$u$} [m/s]& \textbf{$v$} [m/s] & \textbf{$w$} [m/s]& \textbf{$\zeta$} [m]& \textbf{$u$} [m/s]& \textbf{$v$} [m/s]& \textbf{$w$} [m/s]& \textbf{$\zeta$} [m]\\ 
    \midrule

5  & 3.39 (3.08 + 0.31) & 0.888 & 1.80E-02 & 1.73E-02 & 9.60E-05 & 4.58E-02 & 2.89E-02 & 2.61E-02 & 3.57E-04 & 7.25E-02 \\ 
15 & 1.32 (0.72 + 0.60)  & 0.925 & 2.70E-02 & 2.52E-02 & 1.16E-04 & 1.11E-01 & 4.22E-02 & 3.67E-02 & 4.04E-04 & 1.55E-01 \\ 
25 & 1.83 (0.65 + 1.18)  & 0.937 & 2.72E-02 & 2.57E-02 & 1.15E-04 & 1.09E-01 & 4.17E-02 & 3.71E-02 & 3.86E-04 & 1.41E-01 \\ 
\bottomrule
\end{tabular}%
}
\end{table*}

\subsection{Physics-based Verification}

\begin{figure}[t]
\centering
\includegraphics[width=0.4\textwidth]{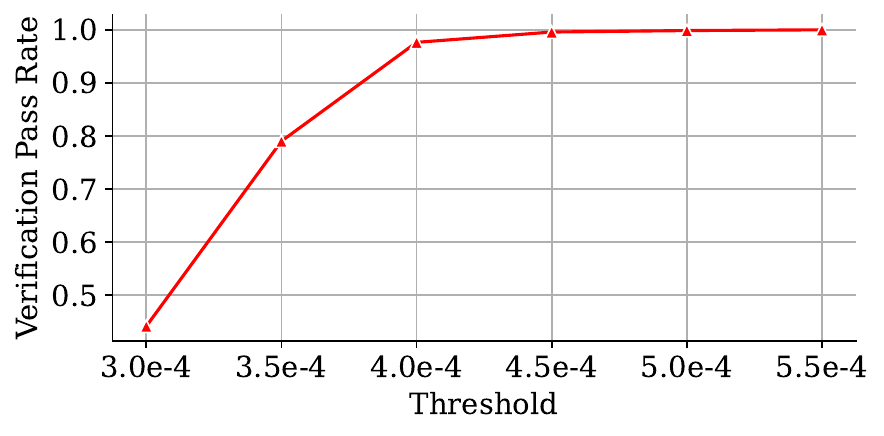}
\caption{ Verification pass rate of AI surrogate inference results at various water mass residual thresholds (m/s). Water mass residual smaller than 5.0e-4 m/s is typically considered acceptable in oceanography. Over 99\% of AI surrogate results fall below this threshold, demonstrating high quality performance. }
\label{fig:threshold}
\end{figure}

We assess the performance of the AI surrogate-based workflow. First, we quantify the AI surrogate inference quality by comparing water mass residuals calculated from AI surrogate results with various thresholds. 
We set the threshold of passing the verification to 3e-4 m/s, 3.5e-4 m/s, 4e-4 m/s, 4.5e-4 m/s,  5e-4 m/s, and 5.5e-4 m/s and tested on the 2012 dataset.
Figure~\ref{fig:threshold} shows these results.
As the water mass residual threshold increases, the acceptance rate of AI surrogate inference results also increases. This is because higher thresholds allow for results that deviate more from strict adherence to the water mass conservation law while still being considered acceptable.
Water mass residuals of simulated results smaller than 5.0e-4 m/s are typically considered acceptable by oceanographers. Our analysis shows that over 99\% of AI surrogate inference results meet this criterion, indicating that they are of high quality.

\begin{figure}
    \centering
\includegraphics[width=0.48\textwidth]{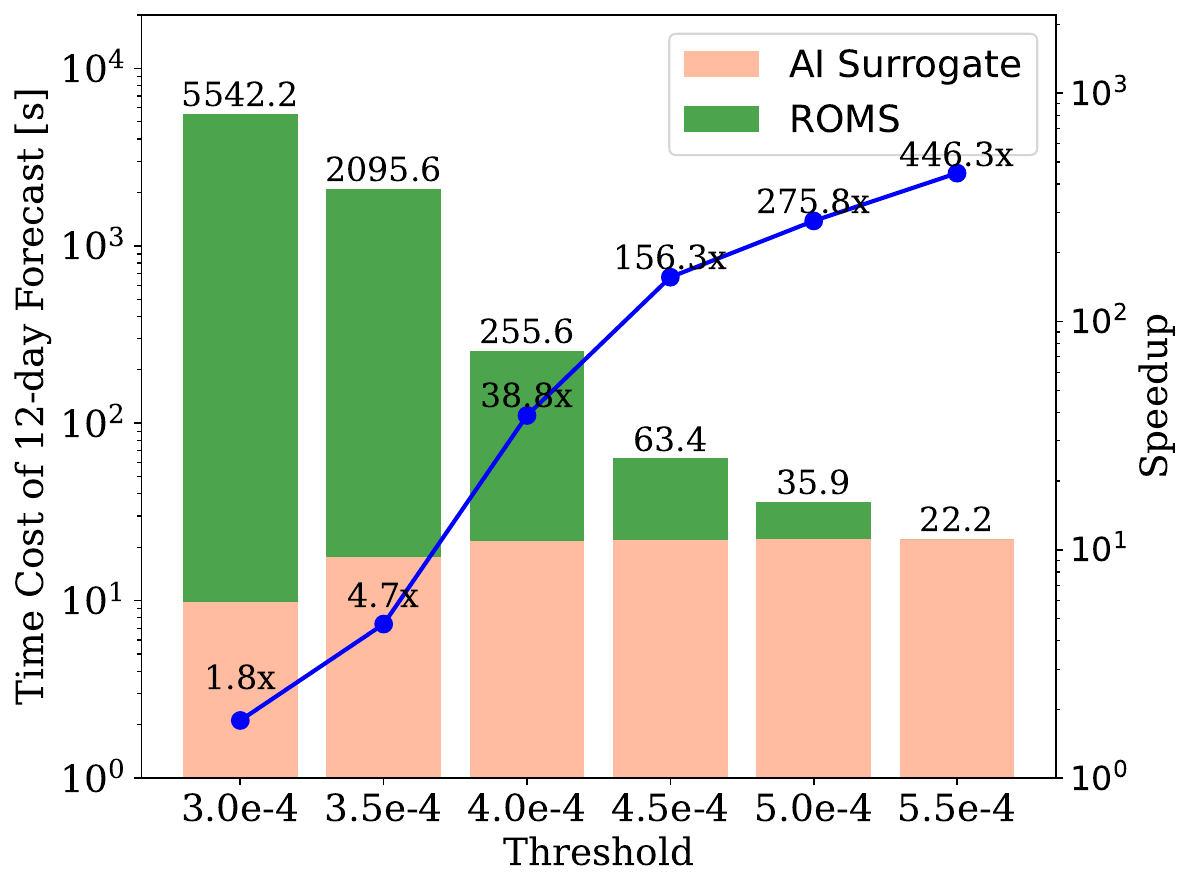}
\caption{End-to-end efficiency of integrated workflow with AI surrogate inference results verification and correction. The x-axis represents various water mass residual thresholds. Bar heights indicate the total execution time of the workflow at different verification thresholds. The blue line shows the speedup factor of the AI surrogate-based workflow compared to the MPI-based ROMS simulation. }
\label{fig:verification}
\end{figure}

We further quantify the end-to-end performance of the AI surrogate-based workflow for coastal ocean circulation simulation. Figure~\ref{fig:verification} illustrates the execution time of the workflow at various verification thresholds, comprising AI surrogate computation time plus ROMS execution time for cases where AI surrogate inference fails verification. The verification time can be ignored.
Specifically,  the workflow switches back to ROMS for computation after detecting low-quality results from the AI surrogate. 
Notably, as discussed in Section~\ref{sec:workflow}, we employ dual-model inference for long-term forecasting. Since the verification is very lightweight, we check inference quality after every AI surrogate inference, enabling early error detection during the calculation. Evidence of this is seen in the varying AI surrogate execution times in Figure~\ref{fig:verification}. 

The AI surrogate-based workflow demonstrates significant efficiency gains, as shown in Figure~\ref{fig:verification}. It achieves up to a 450$\times$ speedup compared to traditional methods. Even with a strict verification threshold (i.e., water mass residual smaller than 3e-4 m/s), the workflow still outperforms traditional ROMS simulation by 1.8$\times$. Furthermore, using the threshold typically accepted by oceanographers, the AI surrogate-based workflow achieves up to a 275$\times$ speedup. These results highlight the substantial computational efficiency gained through the AI surrogate approach while maintaining simulation quality acceptable to the oceanographic community.

\subsection{Sensitivity Analysis}

We conduct a sensitivity analysis in patching size for the AI surrogate efficiency. 
As a key hyperparameter within the AI model architecture, patch size critically impacts the system's computational efficiency and predictive accuracy. We explore the effects of three patch sizes—5 (utilized in our AI surrogate system), 15, and 25 horizontally. The results, as depicted in Table~\ref{tab:sensitivity}, reveal an effect of patch size on the model's complexity, primarily reflected through the parameter count. Larger patch sizes would decrease the number of patches in attention computations, thereby reducing parameters, but will increase the decoder's parameter count because of the transpose convolutional upsampling operations. The smallest patch size assessed, 5, although has the largest parameter count mostly achieved the lowest MAE and RMSE for the 4 variables, underscoring its ability to capture spatiotemporal dynamics with finer resolution. Conversely, enlarging the patch size led to a marginal increase in inference time.

\subsection{Ablation Study}

\begin{figure}[t]
\centering
\includegraphics[width=0.48\textwidth]{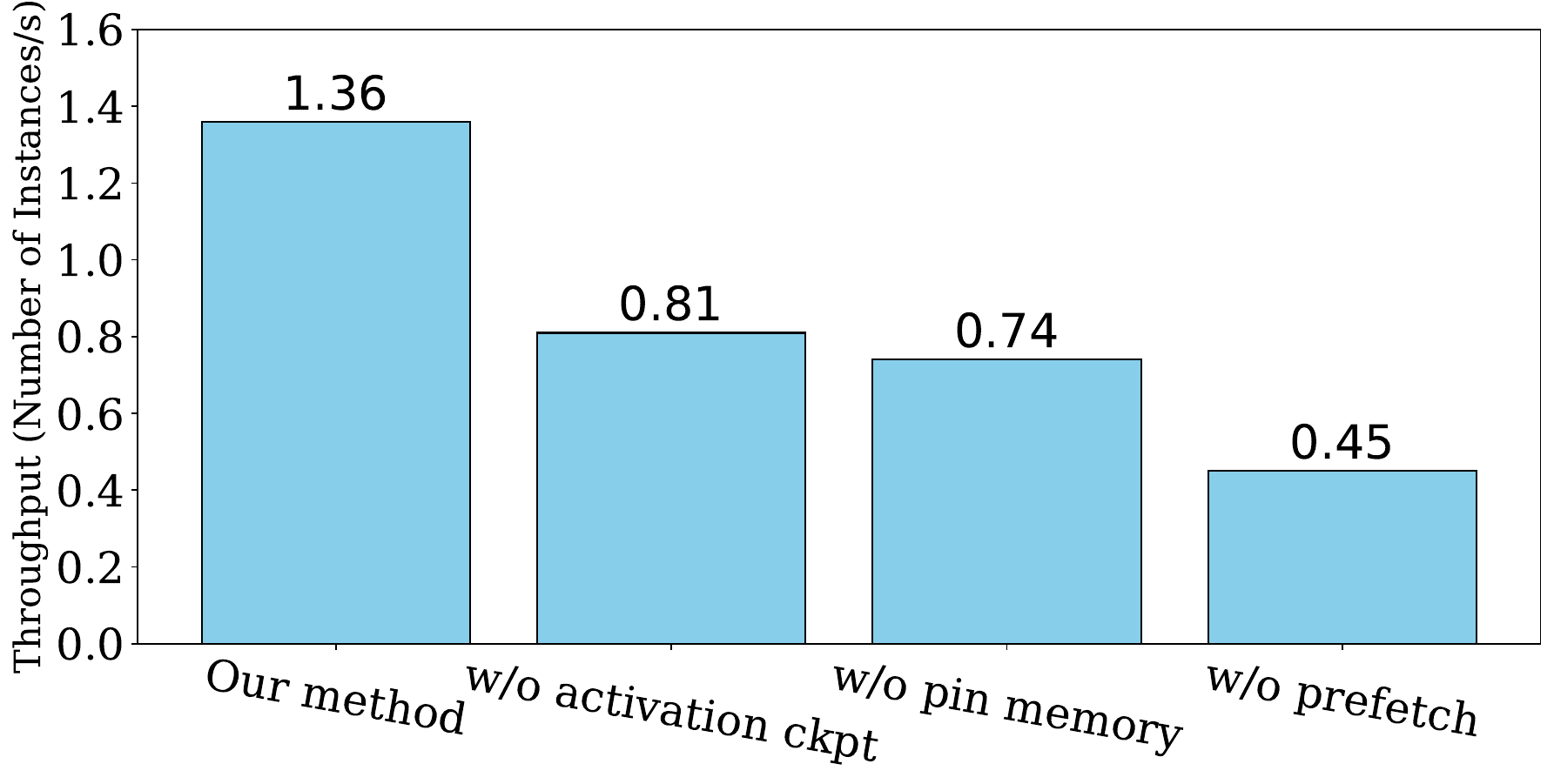}
\caption{Ablation study showing the impact of activation checkpointing, pinned memory, and data prefetch on training throughput. Experiments were conducted with 1 GPU, 16 CPU cores, and 250 GB RAM.}
\label{fig:ablation}
\end{figure}

We conduct an ablation study to evaluate the impact of various system optimization strategies on the AI surrogate training process. 
We test the performance on 1 GPU with 16 CPU cores and 250 GB RAM. Figure~\ref{fig:ablation} illustrates the training throughput associated with our proposed method and its counterparts, each lacking one of the optimizations: pin memory, activation checkpointing, and data prefetching. The analysis indicates that removing pin memory results in the most substantial decrease in throughput, underscoring its critical role in efficient data transfer from CPU to GPU. The omission of activation checkpointing also increases time costs, suggesting their roles in enhancing computational efficiency through effective memory management. During model training, data prefetching is used to load data in advance, this results show that prefetching significantly reduces idle time by overlapping data loading with model computation, thus improving training efficiency.

\subsection{Scalability}

\begin{figure}[t]
\centering
\includegraphics[width=0.4\textwidth]{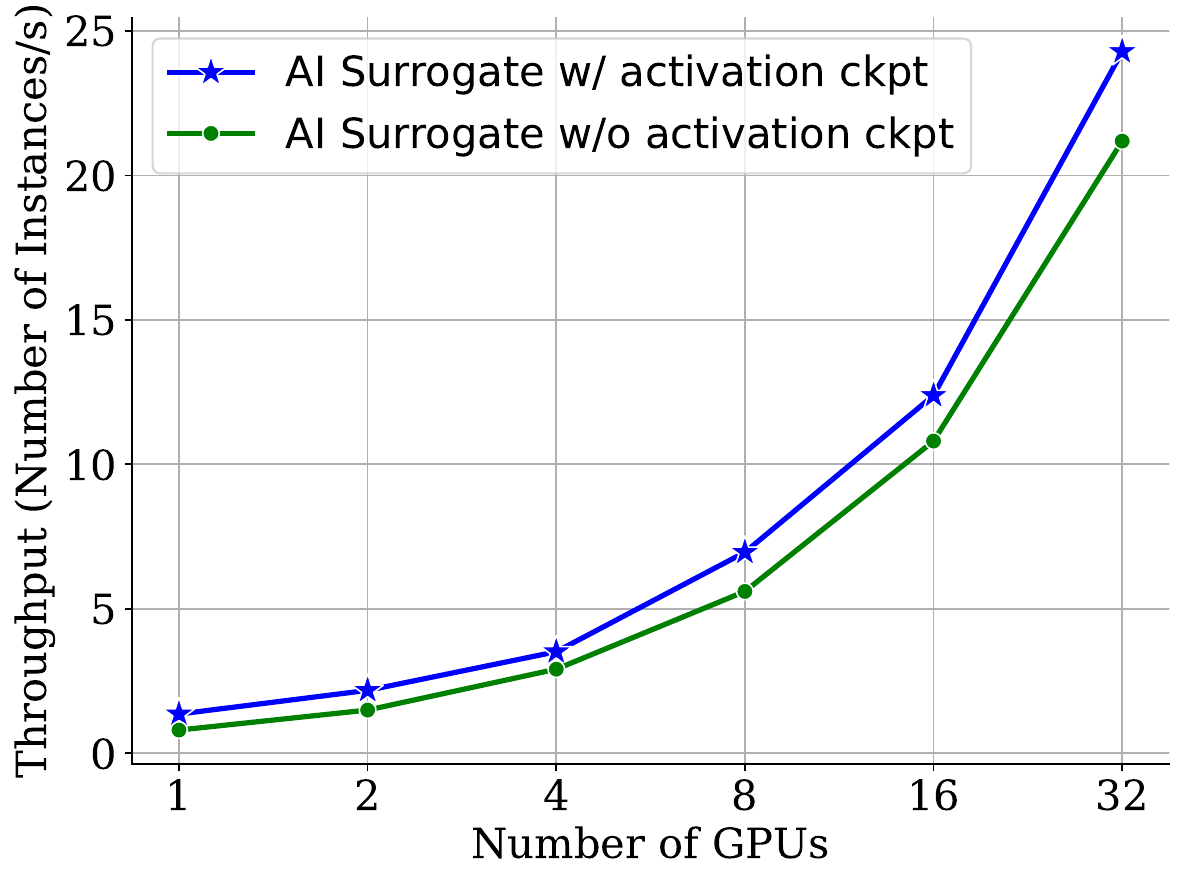}
\caption{Scalability of AI surrogate training with and without activation checkpointing, using 1 to 32 GPUs. Experiments use 1, 2, 4, and 8 GPUs on a single compute node, while 16 and 32 GPU experiments utilize 2 and 4 compute nodes, respectively.}
\label{fig:scaling}
\end{figure}

Figure~\ref{fig:scaling} illustrates the exploration of the weak scalability in the training AI surrogate model, presenting the training throughput as the system proportionally scales with the number of GPUs. 
Due to the large training instance size, an A100 GPU with 80 GB memory can initially train only one instance per GPU. By enabling activation checkpointing, we increase the training batch size to 2 per GPU, achieving higher training throughput by fully utilizing GPU computational capacity.
The implementation of activation checkpointing not only allows for a larger batch size but also demonstrates improved memory efficiency. This enables simultaneous processing of more data, potentially leading to faster convergence and better utilization of available hardware. These improvements pave the way for AI surrogate training with even larger simulation sizes.



\section{Conclusion and Future Work}

This study introduces an AI surrogate model based on a 4D Swin transformer for Regional Ocean Modeling System (ROMS) simulation acceleration. By leveraging GPU's massive parallelism, which traditional ROMS simulations on CPUs cannot utilize, the AI surrogate achieves up to 450× speedup while maintaining high-quality results. 
To ensure reliability, we implement a workflow with physical law-based verification, automatically reverting to traditional simulations if the AI surrogate's solution is deemed unreasonable. 
The proposed workflow enables high-quality, rapid simulations and paves the way for improved real-time forecasting in coastal hazard management, with potential applications in early warning systems and rapid disaster response.

In future work, we plan to extend the model to other coastal processes (e.g., storm surge), explore additional neural network architectures such as diffusion models, and design an uncertainty quantification module for the AI surrogate. We will also explore model parallelism to scale up our AI surrogate to very large meshes.

\section*{Acknowledgments}
This material is based upon work supported by the National Science Foundation (NSF) under Grant No. IIS-2147908, IIS-2207072, OAC-2152085, OAC-2402946, OAC-2410884, and OAC-2402947.

\bibliographystyle{IEEEtran}
\bibliography{reference}

\begin{thebibliography}{10}
\providecommand{\url}[1]{#1}
\csname url@samestyle\endcsname
\providecommand{\newblock}{\relax}
\providecommand{\bibinfo}[2]{#2}
\providecommand{\BIBentrySTDinterwordspacing}{\spaceskip=0pt\relax}
\providecommand{\BIBentryALTinterwordstretchfactor}{4}
\providecommand{\BIBentryALTinterwordspacing}{\spaceskip=\fontdimen2\font plus
\BIBentryALTinterwordstretchfactor\fontdimen3\font minus \fontdimen4\font\relax}
\providecommand{\BIBforeignlanguage}[2]{{%
\expandafter\ifx\csname l@#1\endcsname\relax
\typeout{** WARNING: IEEEtran.bst: No hyphenation pattern has been}%
\typeout{** loaded for the language `#1'. Using the pattern for}%
\typeout{** the default language instead.}%
\else
\language=\csname l@#1\endcsname
\fi
#2}}
\providecommand{\BIBdecl}{\relax}
\BIBdecl

\bibitem{uncoast}
{United Nations}, ``{Climate Change-induced Sea-Level Rise Direct Threat to Millions around World, Secretary-General Tells Security Council},'' \url{https://press.un.org/en/2023/sc15199.doc.htm}, 2023.

\bibitem{gurvan_madec_2023_8167700}
\BIBentryALTinterwordspacing
G.~Madec, M.~Bell, A.~Blaker, C.~Bricaud, D.~Bruciaferri, M.~Castrillo, D.~Calvert, J.~Chanut, E.~Clementi, A.~Coward, I.~Epicoco, C.~Éthé, J.~Ganderton, J.~Harle, K.~Hutchinson, D.~Iovino, D.~Lea, T.~Lovato, M.~Martin, N.~Martin, F.~Mele, D.~Martins, S.~Masson, P.~Mathiot, F.~Mele, S.~Mocavero, S.~Müller, A.~G. Nurser, S.~Paronuzzi, M.~Peltier, R.~Person, C.~Rousset, S.~Rynders, G.~Samson, S.~Téchené, M.~Vancoppenolle, and C.~Wilson, ``Nemo ocean engine reference manual,'' Jul. 2023. [Online]. Available: \url{https://doi.org/10.5281/zenodo.8167700}
\BIBentrySTDinterwordspacing

\bibitem{mitgcm}
MIT, ``{Massachusetts Institute of Technology General Circulation Model},'' \url{https://mitgcm.org/about-mitgcm/}, 2023.

\bibitem{HYCOM}
C.~for Ocean-Atmospheric Prediction Studies~(COAPS), ``{HYbrid Coordinate Ocean Model (HYCOM)},'' \url{https://www.hycom.org/hycom}, 2023.

\bibitem{zhang2016seamless}
Y.~J. Zhang, F.~Ye, E.~V. Stanev, and S.~Grashorn, ``Seamless cross-scale modeling with schism,'' \emph{Ocean Modelling}, vol. 102, pp. 64--81, 2016.

\bibitem{shchepetkin2005regional}
A.~F. Shchepetkin and J.~C. McWilliams, ``The regional oceanic modeling system (roms): a split-explicit, free-surface, topography-following-coordinate oceanic model,'' \emph{Ocean modelling}, vol.~9, no.~4, pp. 347--404, 2005.

\bibitem{roms}
D.~O.~M. Group, ``{Regional Ocean Modeling System (ROMS)},'' \url{https://www.myroms.org/}, 2023.

\bibitem{wang2005parallel}
P.~Wang, Y.~Song, Y.~Chao, and H.~Zhang, ``Parallel computation of the regional ocean modeling system (roms),'' \emph{International Journal of High Performance Computing Applications, vol. 19, no. 4, December 1, 2005, pp. 375-385}, vol.~19, no. UCRL-JRNL-211096, 2005.

\bibitem{panzer2013high}
I.~Panzer, S.~Lines, J.~Mak, P.~Choboter, and C.~Lupo, ``High performance regional ocean modeling with gpu acceleration,'' in \emph{2013 OCEANS-San Diego}.\hskip 1em plus 0.5em minus 0.4em\relax IEEE, 2013, pp. 1--4.

\bibitem{mak2011numerical}
J.~Mak, P.~Choboter, and C.~Lupo, ``Numerical ocean modeling and simulation with cuda,'' in \emph{OCEANS'11 MTS/IEEE KONA}.\hskip 1em plus 0.5em minus 0.4em\relax IEEE, 2011, pp. 1--6.

\bibitem{REMORA}
L.~B.~N. Laboratory, ``{Regional Model of the Ocean Refined Adaptively (REMORA)},'' \url{https://github.com/seahorce-scidac/REMORA}, 2023.

\bibitem{krasnopolsky2006complex}
V.~M. Krasnopolsky and M.~S. Fox-Rabinovitz, ``Complex hybrid models combining deterministic and machine learning components for numerical climate modeling and weather prediction,'' \emph{Neural Networks}, vol.~19, no.~2, pp. 122--134, 2006.

\bibitem{o2018using}
P.~A. O'Gorman and J.~G. Dwyer, ``Using machine learning to parameterize moist convection: Potential for modeling of climate, climate change, and extreme events,'' \emph{Journal of Advances in Modeling Earth Systems}, vol.~10, no.~10, pp. 2548--2563, 2018.

\bibitem{reichstein2019deep}
M.~Reichstein, G.~Camps-Valls, B.~Stevens, M.~Jung, J.~Denzler, N.~Carvalhais \emph{et~al.}, ``Deep learning and process understanding for data-driven earth system science,'' \emph{Nature}, vol. 566, no. 7743, pp. 195--204, 2019.

\bibitem{mohan2018deep}
A.~T. Mohan and D.~V. Gaitonde, ``A deep learning based approach to reduced order modeling for turbulent flow control using lstm neural networks,'' \emph{arXiv preprint arXiv:1804.09269}, 2018.

\bibitem{xiao2019reduced}
D.~Xiao, C.~Heaney, L.~Mottet, F.~Fang, W.~Lin, I.~Navon, Y.~Guo, O.~Matar, A.~Robins, and C.~Pain, ``A reduced order model for turbulent flows in the urban environment using machine learning,'' \emph{Building and Environment}, vol. 148, pp. 323--337, 2019.

\bibitem{bode2021using}
M.~Bode, M.~Gauding, Z.~Lian, D.~Denker, M.~Davidovic, K.~Kleinheinz, J.~Jitsev, and H.~Pitsch, ``Using physics-informed enhanced super-resolution generative adversarial networks for subfilter modeling in turbulent reactive flows,'' \emph{Proceedings of the Combustion Institute}, vol.~38, no.~2, pp. 2617--2625, 2021.

\bibitem{tanaka2021deep}
A.~Tanaka, A.~Tomiya, and K.~Hashimoto, \emph{Deep Learning and Physics}.\hskip 1em plus 0.5em minus 0.4em\relax Springer, 2021.

\bibitem{li2023rapid}
Y.~Li, X.~Ju, Y.~Xiao, Q.~Jia, Y.~Zhou, S.~Qian, R.~Lin, B.~Yang, S.~Shi, X.~Liu \emph{et~al.}, ``Rapid simulations of atmospheric data assimilation of hourly-scale phenomena with modern neural networks,'' in \emph{Proceedings of the International Conference for High Performance Computing, Networking, Storage and Analysis}, 2023, pp. 1--13.

\bibitem{lam2023learning}
R.~Lam, A.~Sanchez-Gonzalez, M.~Willson, P.~Wirnsberger, M.~Fortunato, F.~Alet, S.~Ravuri, T.~Ewalds, Z.~Eaton-Rosen, W.~Hu \emph{et~al.}, ``Learning skillful medium-range global weather forecasting,'' \emph{Science}, vol. 382, no. 6677, pp. 1416--1421, 2023.

\bibitem{fourcastnet}
J.~Pathak, S.~Subramanian, P.~Harrington, S.~Raja, A.~Chattopadhyay, M.~Mardani, T.~Kurth, D.~Hall, Z.~Li, K.~Azizzadenesheli, P.~Hassanzadeh, K.~Kashinath, and A.~Anandkumar, ``{FourCastNet}: A global data-driven high-resolution weather model using adaptive fourier neural operators,'' 2022.

\bibitem{bi2023accurate}
K.~Bi, L.~Xie, H.~Zhang, X.~Chen, X.~Gu, and Q.~Tian, ``Accurate medium-range global weather forecasting with 3d neural networks,'' \emph{Nature}, vol. 619, no. 7970, pp. 533--538, 2023.

\bibitem{jung2021containers}
K.~Jung, Y.-K. Cho, and Y.-J. Tak, ``Containers and orchestration of numerical ocean model for computational reproducibility and portability in public and private clouds: Application of roms 3.6,'' \emph{Simulation Modelling Practice and Theory}, vol. 109, p. 102305, 2021.

\bibitem{Nur_2018}
\BIBentryALTinterwordspacing
A.~A. Nur, I.~Mandang, S.~Mubarrok, and M.~Riza, ``The changes of water mass characteristics using 3-dimensional regional ocean modeling system (roms) in balikpapan bay, indonesia,'' \emph{IOP Conference Series: Earth and Environmental Science}, vol. 162, no.~1, p. 012006, jun 2018. [Online]. Available: \url{https://dx.doi.org/10.1088/1755-1315/162/1/012006}
\BIBentrySTDinterwordspacing

\bibitem{de2021impact}
T.~P. de~Paula, J.~A.~M. Lima, C.~A.~S. Tanajura, M.~Andrioni, R.~P. Martins, and W.~Z. Arruda, ``The impact of ocean data assimilation on the simulation of mesoscale eddies at s{\~a}o paulo plateau (brazil) using the regional ocean modeling system,'' \emph{Ocean Modelling}, vol. 167, p. 101889, 2021.

\bibitem{raissi2019physics}
M.~Raissi, P.~Perdikaris, and G.~E. Karniadakis, ``Physics-informed neural networks: A deep learning framework for solving forward and inverse problems involving nonlinear partial differential equations,'' \emph{Journal of Computational physics}, vol. 378, pp. 686--707, 2019.

\bibitem{feng2023physics}
D.~Feng, Z.~Tan, and Q.~He, ``Physics-informed neural networks of the saint-venant equations for downscaling a large-scale river model,'' \emph{Water Resources Research}, vol.~59, no.~2, p. e2022WR033168, 2023.

\bibitem{chen2022physics}
Q.~Chen, N.~Wang, and Z.~Chen, ``Physics-informed deep learning of nearshore wave processes,'' \emph{Coastal Engineering Proceedings}, no.~37, pp. 14--14, 2022.

\bibitem{cuomo2022scientific}
S.~Cuomo, V.~S. Di~Cola, F.~Giampaolo, G.~Rozza, M.~Raissi, and F.~Piccialli, ``Scientific machine learning through physics--informed neural networks: Where we are and what’s next,'' \emph{Journal of Scientific Computing}, vol.~92, no.~3, p.~88, 2022.

\bibitem{guo2020solving}
Y.~Guo, X.~Cao, B.~Liu, and M.~Gao, ``Solving partial differential equations using deep learning and physical constraints,'' \emph{Applied Sciences}, vol.~10, no.~17, p. 5917, 2020.

\bibitem{kovachki2023neural}
N.~B. Kovachki, Z.~Li, B.~Liu, K.~Azizzadenesheli, K.~Bhattacharya, A.~M. Stuart, and A.~Anandkumar, ``Neural operator: Learning maps between function spaces with applications to pdes.'' \emph{J. Mach. Learn. Res.}, vol.~24, no.~89, pp. 1--97, 2023.

\bibitem{huang2022regional}
L.~Huang, Y.~Jing, H.~Chen, L.~Zhang, and Y.~Liu, ``A regional wind wave prediction surrogate model based on cnn deep learning network,'' \emph{Applied Ocean Research}, vol. 126, p. 103287, 2022.

\bibitem{anandkumar2020neural}
A.~Anandkumar, K.~Azizzadenesheli, K.~Bhattacharya, N.~Kovachki, Z.~Li, B.~Liu, and A.~Stuart, ``Neural operator: Graph kernel network for partial differential equations,'' in \emph{ICLR 2020 Workshop on Integration of Deep Neural Models and Differential Equations}, 2020.

\bibitem{li2020multipole}
Z.~Li, N.~Kovachki, K.~Azizzadenesheli, B.~Liu, A.~Stuart, K.~Bhattacharya, and A.~Anandkumar, ``Multipole graph neural operator for parametric partial differential equations,'' \emph{Advances in Neural Information Processing Systems}, vol.~33, pp. 6755--6766, 2020.

\bibitem{li2020fourier}
Z.~Li, N.~Kovachki, K.~Azizzadenesheli, B.~Liu, K.~Bhattacharya, A.~Stuart, and A.~Anandkumar, ``Fourier neural operator for parametric partial differential equations,'' \emph{arXiv preprint arXiv:2010.08895}, 2020.

\bibitem{guibas2021adaptive}
J.~Guibas, M.~Mardani, Z.~Li, A.~Tao, A.~Anandkumar, and B.~Catanzaro, ``Adaptive fourier neural operators: Efficient token mixers for transformers,'' \emph{arXiv preprint arXiv:2111.13587}, 2021.

\bibitem{wu2022nodeformer}
Q.~Wu, W.~Zhao, Z.~Li, D.~P. Wipf, and J.~Yan, ``Nodeformer: A scalable graph structure learning transformer for node classification,'' \emph{Advances in Neural Information Processing Systems}, vol.~35, pp. 27\,387--27\,401, 2022.

\bibitem{cao2021choose}
S.~Cao, ``Choose a transformer: Fourier or galerkin,'' \emph{Advances in neural information processing systems}, vol.~34, pp. 24\,924--24\,940, 2021.

\bibitem{nguyen2023climax}
T.~Nguyen, J.~Brandstetter, A.~Kapoor, J.~K. Gupta, and A.~Grover, ``Climax: a25 foundation model for weather and climate,'' in \emph{Proceedings of the 40th International Conference on Machine Learning}, 2023, pp. 25\,904--25\,938.

\bibitem{kim2015time}
S.-W. Kim, J.~A. Melby, N.~C. Nadal-Caraballo, and J.~Ratcliff, ``A time-dependent surrogate model for storm surge prediction based on an artificial neural network using high-fidelity synthetic hurricane modeling,'' \emph{Natural Hazards}, vol.~76, pp. 565--585, 2015.

\bibitem{dong2022recent}
C.~Dong, G.~Xu, G.~Han, B.~J. Bethel, W.~Xie, and S.~Zhou, ``Recent developments in artificial intelligence in oceanography,'' \emph{Ocean-Land-Atmosphere Research}, 2022.

\bibitem{pan2023neural}
S.~Pan, S.~L. Brunton, and J.~N. Kutz, ``Neural implicit flow: a mesh-agnostic dimensionality reduction paradigm of spatio-temporal data,'' \emph{Journal of Machine Learning Research}, vol.~24, no.~41, pp. 1--60, 2023.

\bibitem{yin2022continuous}
Y.~Yin, M.~Kirchmeyer, J.-Y. Franceschi, A.~Rakotomamonjy, and P.~Gallinari, ``Continuous {PDE} dynamics forecasting with implicit neural representations,'' in \emph{The Eleventh International Conference on Learning Representations}, 2022.

\bibitem{chen2022crom}
P.~Y. Chen, J.~Xiang, D.~H. Cho, Y.~Chang, G.~Pershing, H.~T. Maia, M.~Chiaramonte, K.~Carlberg, and E.~Grinspun, ``{CROM}: Continuous reduced-order modeling of {PDEs} using implicit neural representations,'' \emph{arXiv preprint arXiv:2206.02607}, 2022.

\bibitem{li2022fourier}
Z.~Li, D.~Z. Huang, B.~Liu, and A.~Anandkumar, ``Fourier neural operator with learned deformations for {PDEs} on general geometries,'' \emph{arXiv preprint arXiv:2207.05209}, 2022.

\bibitem{li2023geometry}
Z.~Li, N.~B. Kovachki, C.~Choy, B.~Li, J.~Kossaifi, S.~P. Otta, M.~A. Nabian, M.~Stadler, C.~Hundt, K.~Azizzadenesheli, and A.~Anandkumar, ``Geometry-informed neural operator for large-scale {3D PDEs},'' \emph{arXiv preprint arXiv:2309.00583}, 2023.

\bibitem{vinuesa2022enhancing}
R.~Vinuesa and S.~L. Brunton, ``Enhancing computational fluid dynamics with machine learning,'' \emph{Nature Computational Science}, vol.~2, no.~6, pp. 358--366, 2022.

\bibitem{li2022fourier2}
Z.~Li, W.~Peng, Z.~Yuan, and J.~Wang, ``Fourier neural operator approach to large eddy simulation of three-dimensional turbulence,'' \emph{Theoretical and Applied Mechanics Letters}, vol.~12, no.~6, p. 100389, 2022.

\bibitem{taghizadeh2021turbulence}
S.~Taghizadeh, F.~D. Witherden, Y.~A. Hassan, and S.~S. Girimaji, ``Turbulence closure modeling with data-driven techniques: Investigation of generalizable deep neural networks,'' \emph{Physics of Fluids}, vol.~33, no.~11, 2021.

\bibitem{gupta2022three}
R.~Gupta and R.~Jaiman, ``Three-dimensional deep learning-based reduced order model for unsteady flow dynamics with variable reynolds number,'' \emph{Physics of Fluids}, vol.~34, no.~3, 2022.

\bibitem{wang2020reduced}
M.~Wang, S.~W. Cheung, W.~T. Leung, E.~T. Chung, Y.~Efendiev, and M.~Wheeler, ``Reduced-order deep learning for flow dynamics. the interplay between deep learning and model reduction,'' \emph{Journal of Computational Physics}, vol. 401, p. 108939, 2020.

\bibitem{warner2008development}
J.~C. Warner, C.~R. Sherwood, R.~P. Signell, C.~K. Harris, and H.~G. Arango, ``Development of a three-dimensional, regional, coupled wave, current, and sediment-transport model,'' \emph{Computers \& geosciences}, vol.~34, no.~10, pp. 1284--1306, 2008.

\bibitem{warner2010development}
J.~C. Warner, B.~Armstrong, R.~He, and J.~B. Zambon, ``Development of a coupled ocean--atmosphere--wave--sediment transport (coawst) modeling system,'' \emph{Ocean modelling}, vol.~35, no.~3, pp. 230--244, 2010.

\bibitem{hsu2017parametric}
C.-E. Hsu, S.-C. Hsiao, and J.-T. Hsu, ``Parametric analyses of wave-induced nearshore current system,'' \emph{Journal of Coastal Research}, vol.~33, no.~4, pp. 795--801, 2017.

\bibitem{liu2021Swin}
Z.~Liu, Y.~Lin, Y.~Cao, H.~Hu, Y.~Wei, Z.~Zhang, S.~Lin, and B.~Guo, ``Swin transformer: Hierarchical vision transformer using shifted windows,'' in \emph{Proceedings of the IEEE/CVF international conference on computer vision}, 2021, pp. 10\,012--10\,022.

\bibitem{liu2022video}
Z.~Liu, J.~Ning, Y.~Cao, Y.~Wei, Z.~Zhang, S.~Lin, and H.~Hu, ``Video swin transformer,'' in \emph{Proceedings of the IEEE/CVF conference on computer vision and pattern recognition}, 2022, pp. 3202--3211.

\bibitem{kim2024swift}
P.~Kim, J.~Kwon, S.~Joo, S.~Bae, D.~Lee, Y.~Jung, S.~Yoo, J.~Cha, and T.~Moon, ``Swift: Swin 4d fmri transformer,'' \emph{Advances in Neural Information Processing Systems}, vol.~36, 2024.

\bibitem{dosovitskiy2020image}
A.~Dosovitskiy, L.~Beyer, A.~Kolesnikov, D.~Weissenborn, X.~Zhai, T.~Unterthiner, M.~Dehghani, M.~Minderer, G.~Heigold, S.~Gelly \emph{et~al.}, ``An image is worth 16x16 words: Transformers for image recognition at scale,'' in \emph{International Conference on Learning Representations}, 2020.

\bibitem{ioffe2015batch}
S.~Ioffe and C.~Szegedy, ``Batch normalization: Accelerating deep network training by reducing internal covariate shift,'' in \emph{International conference on machine learning}.\hskip 1em plus 0.5em minus 0.4em\relax pmlr, 2015, pp. 448--456.

\bibitem{hendrycks2016gaussian}
D.~Hendrycks and K.~Gimpel, ``Gaussian error linear units (gelus),'' \emph{arXiv preprint arXiv:1606.08415}, 2016.

\bibitem{ronneberger2015u}
O.~Ronneberger, P.~Fischer, and T.~Brox, ``U-net: Convolutional networks for biomedical image segmentation,'' in \emph{Medical image computing and computer-assisted intervention--MICCAI 2015: 18th international conference, Munich, Germany, October 5-9, 2015, proceedings, part III 18}.\hskip 1em plus 0.5em minus 0.4em\relax Springer, 2015, pp. 234--241.

\bibitem{hatamizadeh2021swin}
A.~Hatamizadeh, V.~Nath, Y.~Tang, D.~Yang, H.~R. Roth, and D.~Xu, ``Swin unetr: Swin transformers for semantic segmentation of brain tumors in mri images,'' in \emph{International MICCAI brainlesion workshop}.\hskip 1em plus 0.5em minus 0.4em\relax Springer, 2021, pp. 272--284.

\bibitem{vaswani2017attention}
A.~Vaswani, N.~Shazeer, N.~Parmar, J.~Uszkoreit, L.~Jones, A.~N. Gomez, {\L}.~Kaiser, and I.~Polosukhin, ``Attention is all you need,'' \emph{Advances in neural information processing systems}, vol.~30, 2017.

\bibitem{ba2016layer}
J.~L. Ba, J.~R. Kiros, and G.~E. Hinton, ``Layer normalization,'' \emph{arXiv preprint arXiv:1607.06450}, 2016.

\bibitem{tang2022self}
Y.~Tang, D.~Yang, W.~Li, H.~R. Roth, B.~Landman, D.~Xu, V.~Nath, and A.~Hatamizadeh, ``Self-supervised pre-training of swin transformers for 3d medical image analysis,'' in \emph{Proceedings of the IEEE/CVF conference on computer vision and pattern recognition}, 2022, pp. 20\,730--20\,740.

\bibitem{bertasius2021space}
G.~Bertasius, H.~Wang, and L.~Torresani, ``Is space-time attention all you need for video understanding?'' in \emph{ICML}, vol.~2, no.~3, 2021, p.~4.

\end{thebibliography}

\end{document}